\documentclass[runningheads]{llncs}

\usepackage{eccv}

\usepackage{marvosym}

\usepackage{eccvabbrv}

\usepackage{graphicx}
\usepackage{booktabs}
\usepackage{array}
\usepackage{float}
\usepackage{multirow}

\usepackage[most]{tcolorbox}
\usepackage{listings}
\usepackage{xcolor}
\usepackage{wrapfig}

\usepackage{enumitem}

\usepackage{tikz}
\usepackage{pgfplots}
\pgfplotsset{compat=1.18}

\usepackage[breaklinks,colorlinks,citecolor=eccvblue]{hyperref}
\hypersetup{
  pdftitle={OccDirector: Language-Guided Behavior and Interaction Generation in 4D Occupancy Space},
  pdfauthor={Zhuding Liang, Tianyi Yan, Dubing Chen, Jiasen Zheng, Huan Zheng, Cheng-zhong Xu, Yida Wang, Kun Zhan, Jianbing Shen}
}

\newcommand{\ModelName}{OccDirector}
\newcommand{\DatasetName}{OccInteract-85k}
\newcommand{\StaticDatasetName}{OccInteract-Static}
\newcommand{\EnvDatasetName}{OccInteract-Env}
\newcommand{\AgentDatasetName}{OccInteract-Agent}

\newlength{\figcapgap}     
\newlength{\figbottomgap}  
\newlength{\tabcapgap}     
\newlength{\tabbottomgap}  
\begin{document}


\title{\ModelName{}: Language-Guided Behavior and Interaction Generation in 4D Occupancy Space}

\titlerunning{\ModelName{}: Language-Guided 4D Occupancy Generation}

\author{Zhuding Liang\inst{1}\thanks{Equal Contribution.} \and
Tianyi Yan\inst{1}\textsuperscript{*} \and
Dubing Chen\inst{1} \and
Jiasen Zheng\inst{2} \and
Huan Zheng\inst{1} \and
Cheng-zhong Xu\inst{1} \and
Yida Wang\inst{3} \and
Kun Zhan\inst{3} \and
Jianbing Shen\inst{1,}{\textsuperscript{\Letter}}
}

\authorrunning{Z.~Liang, T.~Yan et al.}

\institute{SKL-IOTSC, CIS, University of Macau, Macau, China \and Northwestern University, Evanston, United States\and
Li Auto Inc, Beijing, China}

\setcounter{tocdepth}{2}%
\let\oldaddcontentsline\addcontentsline
\renewcommand{\addcontentsline}[3]{}

\maketitle

\pdfbookmark[0]{OccDirector: Language-Guided Behavior and Interaction Generation in 4D Occupancy Space}{main.title}

\begingroup
\let\thefootnote\relax
\footnotetext{\Letter \ Corresponding author: \textit{Jianbing Shen}. This work was supported by the Science and Technology Development Fund of Macau SAR (FDCT) under grants 0134/2025/RIA2 and 0102/2023/RIA2, and the Jiangyin Hi-tech Industrial Development Zone under the Taihu Innovation Scheme (EF2025-00003-SKL-IOTSC).}
\endgroup

\pdfbookmark[1]{Abstract}{main.sec.abstract}
\begin{abstract}
  Generative world models increasingly rely on 4D occupancy for realistic
  autonomous driving simulation. However, existing generation frameworks
  are limited in two respects: geometric-condition-based methods depend on
  explicit trajectories and cannot generalize to free-form instructions,
  while text-based methods operate at a coarse attribute level and fail to
  orchestrate complex, sequential multi-agent interactions.
  We propose \ModelName{}, a framework that generates 4D occupancy dynamics
  conditioned solely on natural language scripts, without requiring any
  geometric priors.
  \ModelName{} encodes free-form scripts via a frozen VLM and processes
  occupancy tokens through a Spatio-Temporal MMDiT backbone.
  A history-prefix anchoring strategy further ensures that generated
  sequences remain consistent with any provided context over long horizons.
  To support training and evaluation, we introduce \DatasetName{}, a dataset
  of 85k clips annotated with multi-level language instructions spanning
  static layouts to intricate multi-agent behaviors, alongside a novel
  VLM-based evaluation benchmark.
  Extensive experiments demonstrate state-of-the-art generation quality
  and strong instruction-following fidelity across diverse driving scenarios.
\end{abstract}


\pdfbookmark[1]{1 Introduction}{main.sec.intro}
\section{Introduction}
\label{sec:intro}

\begin{figure}[t]
    \centering
    \includegraphics[width=\linewidth]{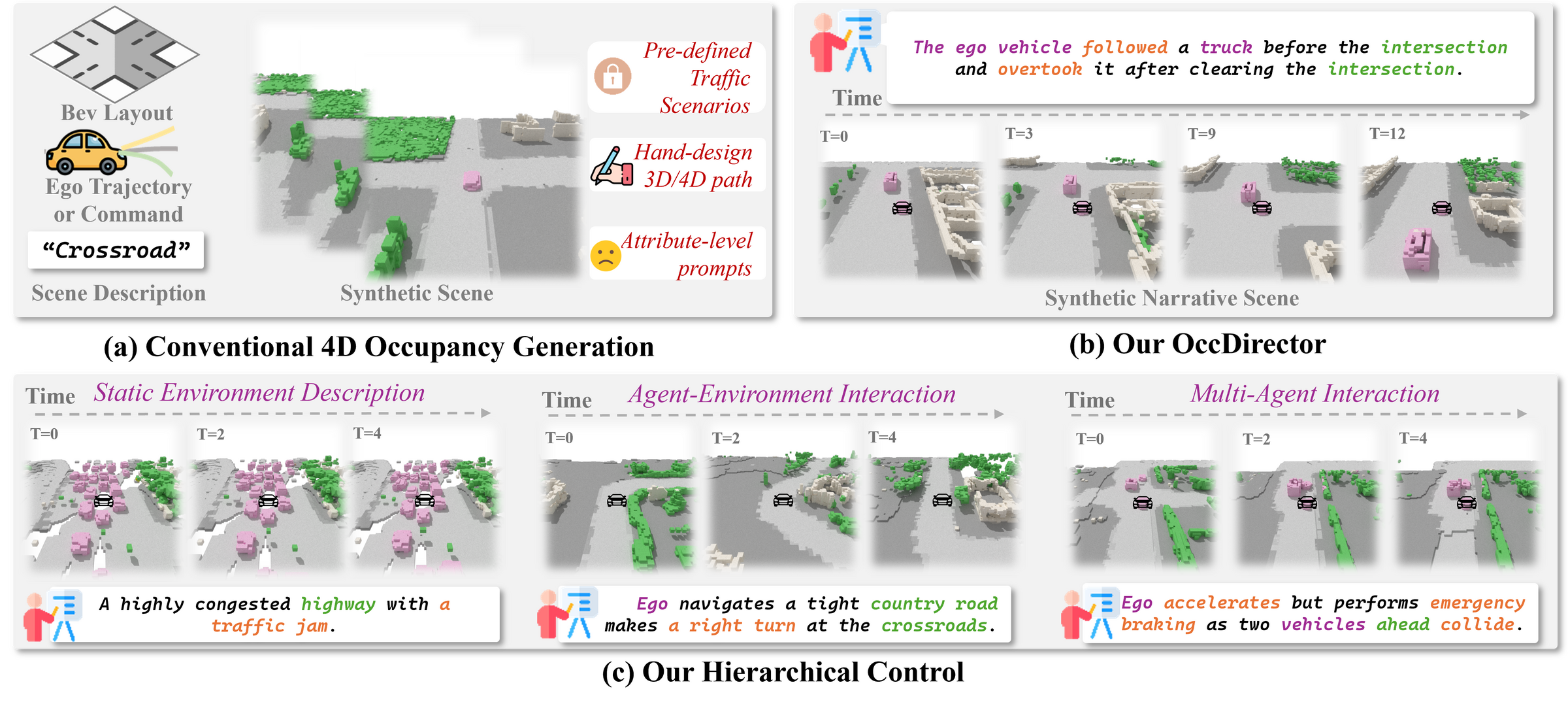}
    \vspace{\figcapgap}
    \caption{\textbf{Script-driven behavior and interaction generation with \ModelName{}.} 
    (a) vs. (b): Comparison between previous methods and our script-driven approach. 
    (c) Hierarchical generation capabilities across Static, Agent-Environment, and Multi-Agent levels.
    The ego vehicle is marked with a \textit{black car icon}. 
    \textit{Note: Due to raw data limitations \cref{sec:data_sources}, some results may exhibit missing ego voxels or sensor blind spots.}
    }
    \label{fig:teaser_controllable_scene_generation}
    \vspace{\figbottomgap}
\end{figure}

The evolution of autonomous driving (AD) systems \cite{li2024data,chen2024end,dong2025end} is increasingly relying on Generative World Models \cite{gaia-2__russell2025gaia,drivedreamer-2__zhao2025drivedreamer,magicdrive-v2__gao2025magicdrive} that can simulate diverse and realistic driving scenarios for closed-loop training and validation. Among various scene representations, 4D Occupancy (space-time occupancy)\cite{khurana2023point,ma2024cam4docc} has emerged as the most unified and comprehensive representation. Unlike 3D bounding boxes which oversimplify object geometry \cite{occnet__tong2023scene,tian2023occ3d}, or LiDAR point clouds \cite{alaba2022survey,wang2023openoccupancy} which suffer from sparsity and occlusion, 4D occupancy grids offer a dense, geometry-aware, and temporally consistent description of dynamic environments. Consequently, synthesizing high-fidelity 4D occupancy rollouts has become a critical frontier in the computer vision community \cite{dio__diehl2025dio,geniedrive__yang2025geniedrive}.

Despite impressive progress in 3D/4D generation, existing occupancy generation frameworks \cite{uniscene__li2025uniscene,occsora__wang2024occsora,dynamiccity__bian2024dynamiccity} heavily rely on explicit geometric conditions—such as BEV semantic layouts, discrete navigation command, or predefined ego-vehicle trajectories—to control the generative process.
As illustrated in \cref{fig:teaser_controllable_scene_generation}(a), methods like \cite{dome__gu2024dome,come__shi2025come,dynamiccity__bian2024dynamiccity} depend on rigid explicit trajectories or simplistic attribute prompts. While offering precise spatial constraints, these conditions are fundamentally inflexible and unscalable. Users must laboriously hand-design accurate 3D paths, a process that cannot intuitively express high-level intents. 
In contrast, our \ModelName{} (\cref{fig:teaser_controllable_scene_generation}(b,c)) acts as a ``scenario director'', interpreting complex natural language scripts to orchestrate dynamic driving scenes, such as \textit{``congested traffic jam''} or \textit{``emergency braking''}.
To date, generating dynamic 4D occupancy purely guided by natural language remains an unsolved challenge. 

Pioneering text-driven 4D occupancy generation faces a significant challenge: the Semantic-Spatiotemporal Gap. Language instructions inherently rely on relational structures and sequential logic to describe abstract behaviors. For instance, consider the instruction: \textit{``The ego vehicle followed a truck before the intersection and overtook it after clearing the intersection.''} Translating such high-level temporal logic into occupancy generation is non-trivial, as it demands rigorous spatio-temporal consistency and physically plausible evolution at the voxel level. However, most existing autonomous driving diffusion models \cite{drivedreamer-1__wang2024drivedreamer,drivedreamer-2__zhao2025drivedreamer,magicdrive__gao2023magicdrive} fail to bridge this gap. Relying on generic text encoders \cite{clip_radford2021learning,t5__raffel2020exploring} that function largely as "bag-of-words" extractors, these models lack the compositional reasoning required to parse complex interactions \cite{yuksekgonul2022and,koishigarina2025clip}. Consequently, they often produce implausible collisions, discontinuous motions, or topological violations when processing intricate instructions  \cite{liang2025worldlens}.

To this end, we introduce \ModelName{}, taking a decisive step toward language-driven scenario orchestration. 
Unlike prior works constrained by geometric priors, \ModelName{} establishes a new paradigm of purely text-guided 4D occupancy generation, bridging the gap from appearance synthesis to complex agent behavior and interaction generation. By replacing rigid trajectory inputs with a director-like natural language interface, OccDirector enables comprehensive controllability along three complementary axes, as visualized in \cref{fig:teaser_controllable_scene_generation}(c): 
(i) \textbf{Static Environment}, synthesizing diverse topologies like \textit{``traffic jams''} via semantic composition; 
(ii) \textbf{Agent-Environment Interaction}, executing topology-constrained maneuvers such as yielding at intersections; and 
(iii) \textbf{Multi-Agent Interaction}, generating physically plausible reactive behaviors like \textit{``emergency braking to avoid collisions''}. 
Technically, OccDirector leverages a powerful Vision-Language Model (VLM) as the text encoder to deeply comprehend procedural semantics. The joint text-occupancy sequence is modeled using a scalable spatio-temporal MMDiT equipped with Spatio-Temporal Separated Attention (STSA) and a history-prefix token-replace strategy for interaction-consistent rollouts. Furthermore, to overcome the complete absence of paired text-4D occupancy data, we develop an automated data engine that translates real-world logs and simulator data into abundant, logic-rich occupancy-language pairs. 

\noindent Our contributions can be summarized as follows:

\begin{itemize}[leftmargin=*, noitemsep, topsep=0pt]
    \item We propose \ModelName{}, a pioneering framework for text-guided 4D occupancy generation, enabling director-like control over agent behaviors without rigid geometric priors.
    
    \item We design a VLM-driven Spatio-Temporal MMDiT with STSA and a history-prefix anchoring strategy, effectively bridging the semantic-spatiotemporal gap for consistent interactions.
    
    \item We introduce \DatasetName{}, a large-scale dataset with hierarchical language instructions ranging from static layouts to complex interactions, alongside a novel VLM-based benchmark.
    
    \item Extensive experiments demonstrate that \ModelName{} achieves state-of-the-art generation quality and unprecedented instruction-following capabilities.
\end{itemize}

\pdfbookmark[1]{2 Related Work}{main.sec.related}
\section{Related Work}

\subsection{Generative World Models for Autonomous Driving}
Generative world models have emerged as a paradigm shift in autonomous driving simulation, enabling the synthesis of diverse driving scenarios for closed-loop training \cite{gaia-1__hu2023gaia,gaia-2__russell2025gaia,drivedreamer-1__wang2024drivedreamer,drivedreamer-2__zhao2025drivedreamer}.
Early approaches predominantly focused on 2D Video Generation \cite{magicdrive__gao2023magicdrive,vista__gao2024vista,kim2021drivegan}, leveraging diffusion models to produce photorealistic camera views. While visually compelling, these perspective-view methods often lack explicit 3D geometric consistency and struggle to support downstream planning tasks that require precise spatial reasoning.

Recently, 4D Occupancy has been established as a unified representation bridging visual fidelity and geometric explicitness \cite{occ-survey1__xu2025survey,occnet__tong2023scene}. By representing the world as a dense, voxelized spatiotemporal volume, it supports physically grounded simulation. However, existing occupancy world models \cite{occworld__zheng2024occworld,occ-transformer__zhang2023occformer} primarily focus on \textit{predicting} future states from past observations (forecasting). They lack the generative flexibility to create new scenarios from scratch based on user intent, limiting their utility for controllable simulation.

\subsection{Conditioned 4D Occupancy Generation}
Unlike forecasting, occupancy generation aims to synthesize plausible 4D scenes conditioned on specific inputs. 
Current state-of-the-art frameworks heavily rely on Explicit Geometric Conditions. For instance, methods like OccSora \cite{occsora__wang2024occsora} and DynamicCity \cite{dynamiccity__bian2024dynamiccity} condition the diffusion process on HD maps, 3D semantic layouts, or pre-defined agent trajectories. 
While these strong priors ensure structural validity, they impose a rigid generation paradigm: users must provide precise, low-level geometric inputs (e.g., drawing a spline for a lane change) to control the output. This ``trajectory-engineering'' process is labor-intensive and non-intuitive.
In contrast, \ModelName{} proposes a paradigm shift from geometry-driven to language-driven generation. We eliminate the need for pre-defined trajectories, allowing users to orchestrate scene evolution through high-level semantic scripts (e.g., \textit{``an aggressive cut-in''}), thereby offering a more scalable interface for scenario design.

\subsection{Text-Conditioned Scene Generation and Fine-Grained Control}
The integration of natural language into generative models has significantly advanced image and video synthesis \cite{Imagen__saharia2022photorealistic,metamorph__tong2025metamorph,sora__liu2024sora,wan__wan2025wan}. In the autonomous driving domain, however, language control remains largely limited to high-level attributes. 
Existing works \cite{drivedreamer-2__zhao2025drivedreamer,magicdrive-v2__gao2025magicdrive,mei2024dreamforge,chen2024unimlvg} typically utilize text prompts to control Global Attributes (e.g., weather conditions, time-of-day) or static scene composition. 
While some approaches like UniScene \cite{uniscene__li2025uniscene} incorporate text, they primarily use it for texture rendering or style transfer, leaving the core spatiotemporal dynamics (i.e., how agents move) determined by other modalities.
Other works like X-Scene \cite{X-scene__yang2025x} adopt a hierarchical approach, converting text into intermediate layouts or bounding boxes before generation. Although this bridges language and static layout, it may struggle to capture the continuous nuances of dynamic interactions.

In this work, we introduce \ModelName{}, which pioneers end-to-end text-to-4D occupancy generation. Unlike methods that treat text as a style modifier or rely on intermediate layout proxies, we learn a direct mapping from procedural language instructions to continuous spatiotemporal voxel flows, enabling fine-grained control over multi-agent interactions and maneuvers purely from natural language.

\pdfbookmark[1]{3 OccDirector: Language-Conditioned 4D Occupancy Interaction Generation}{main.sec.method}
\begin{figure}[t]
    \centering
    \includegraphics[width=\linewidth]{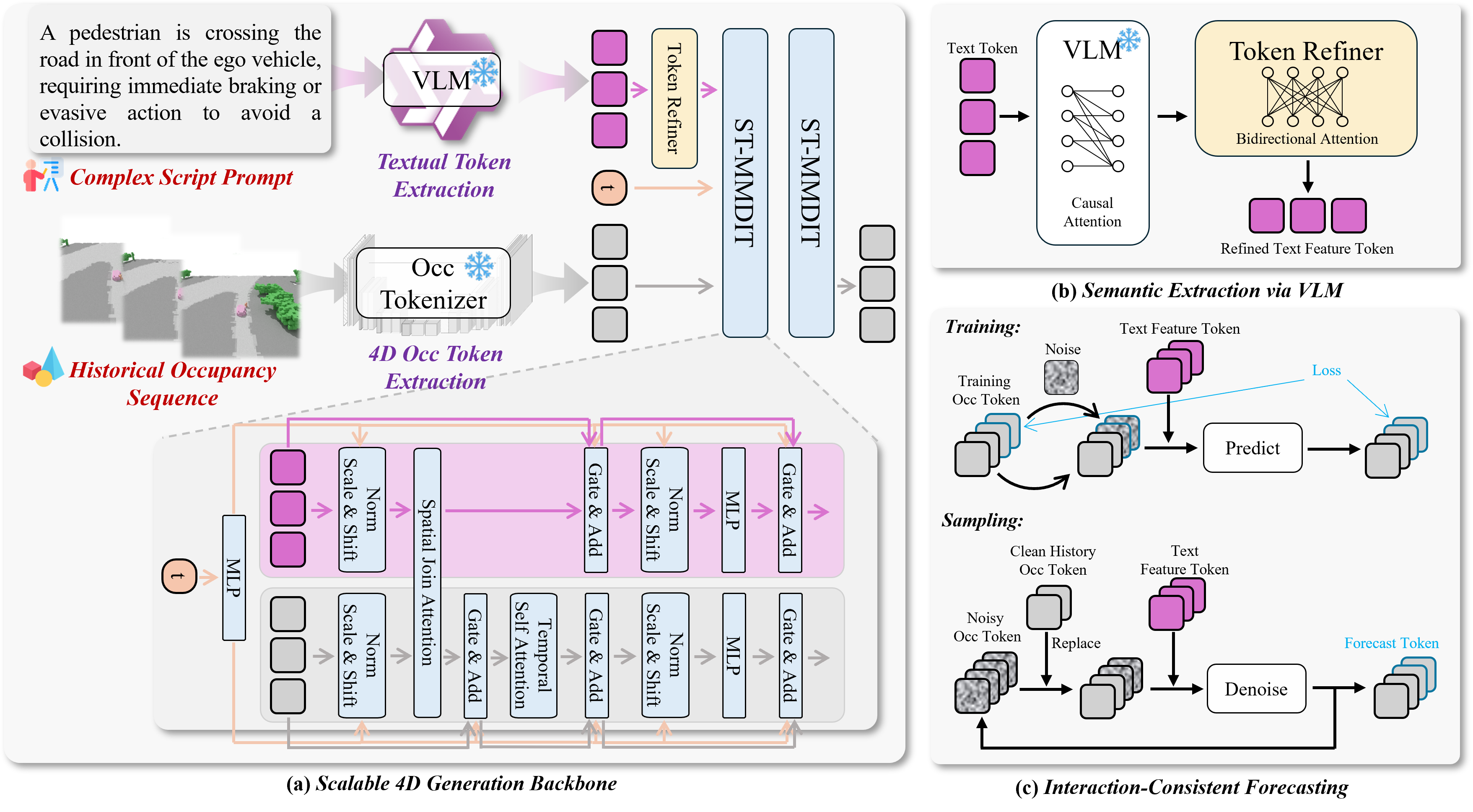}
    \vspace{\figcapgap}
    \caption{
    \textbf{Model Architecture of \ModelName{}.}
    \textbf{(a) Scalable 4D Generation Backbone:} A Spatio-Temporal MMDiT processes text and occupancy tokens via dual streams, utilizing separated spatial and temporal attention. 
    \textbf{(b) Semantic Extraction:} A frozen VLM and a Token Refiner bridge the semantic-spatial gap. 
    \textbf{(c) Interaction-Consistent Forecasting:} The history-anchoring strategy ensures the generated future strictly adheres to the context during both training and sampling.
    }
    \label{fig:model_architecture}
    \vspace{\figbottomgap}
\end{figure}

\section{\ModelName{}: Language-Conditioned 4D Occupancy Interaction Generation}
\label{sec:method}

\subsection{Method Overview} 
\ModelName{} is designed for \emph{script-driven behavior and interaction generation} within 4D occupancy rollouts (\cref{sec:intro}). 
Formally, given a natural language script $\pi$ and an optional history context $\mathbf{O}_{1:h}$, our goal is to generate a future 4D occupancy sequence $\mathbf{O}_{h+1:F}$ that strictly adheres to the spatial layout and dynamic interaction intents specified in $\pi$.
\cref{fig:model_architecture} illustrates our architecture.

\subsection{Joint Latent Representation} \label{sec:encoding}
Directly modeling raw 4D occupancy data and complex natural language is computationally intractable and semantically challenging. To map both modalities into a unified, efficient latent space, we formulate our input encoding stage with three specialized components.

\textbf{Geometry-Aware Occupancy Tokenization.} To alleviate the extreme memory and computational footprint of 4D occupancy sequences $\mathbf{O}_{1:F}\in\{0,\dots,K-1\}^{F\times H\times W\times D}$, we require a geometry-aware compression mechanism. Building upon the discrete auto-encoding principles of DOME \cite{dome__gu2024dome}, we adopt a frozen Occ-VAE. This module maps the discrete 4D rollout into a compact, continuous latent grid, which is subsequently patchified and projected into a sequence of tokens $\mathbf{H}\in\mathbb{R}^{F\times S\times d_{\mathrm{model}}}$, where $F$ is the temporal length and $S$ denotes the number of 3D spatial tokens per frame. Unlike DOME's implementation, we pretrain the Occ-VAE with variable-length rollouts, enabling flexible generation horizons.

\textbf{Procedural Semantic Extraction via VLM.} Effective script-driven generation requires controlling dynamic \emph{interaction intents} (e.g., yielding, cut-in) over time. Standard CLIP-style embeddings \cite{clip_radford2021learning,t5__raffel2020exploring} often lack the relational and procedural semantics needed for such fine-grained control. Recent advancements have demonstrated that leveraging LLMs or VLMs for text feature embedding yields superior performance compared to CLIP-style embeddings \cite{metamorph__tong2025metamorph,janus__wu2025janus,mmada__yang2025mmada,qwen-image__wu2025qwen}. To capture these complex global relations, as shown in \cref{fig:model_architecture}(b), we utilize a frozen Vision-Language Model (VLM) text encoder (Qwen3-VL-2B \cite{qwen3__yang2025qwen3}). Given a prompt $\pi$, the encoder extracts dense, high-dimensional representations $\tilde{\mathbf{C}}\in\mathbb{R}^{L\times d_{\mathrm{text}}}$, where $L$ is the sequence length.

\textbf{Token Refiner.} While VLM features provide rich semantics, they are inherently suboptimal for direct generative control in 4D spatial-temporal layouts due to modality gaps. To bridge this semantic-spatial divide, we introduce a lightweight bidirectional token refiner (\cref{fig:model_architecture}(b)) \cite{hunyuanvideo__kong2024hunyuanvideo,hunyuanvideo-token-refiner__ma2024exploring}. This module, denoted as $\mathrm{Refine}(\cdot)$, employs a shallow transformer alternating between full self-attention and MLP blocks. It projects the raw features to the model dimension, $\mathbf{C}=\mathrm{Refine}(\tilde{\mathbf{C}})\in\mathbb{R}^{L\times d_{\mathrm{model}}}$, producing text tokens with enhanced compositional alignment for occupancy generation.

\subsection{Scalable 4D Generation Backbone}\label{sec:backbone}
Modeling environment constraints and multi-agent interactions in 4D space demands capturing long-range dependencies; failure to do so results in accumulated errors, such as geometry violations or implausible discontinuities. However, the cubic complexity of 3D spatial tokens ($S$) multiplied by the temporal dimension ($F$) makes standard full attention computationally prohibitive.

To resolve this scalability bottleneck while maintaining strict text-to-layout alignment, we formulate a double-stream transformer backbone (\cref{fig:model_architecture}(a)) that builds upon the MMDiT architecture \cite{mmdit-sd3__esser2024scaling}. This design preserves a dedicated text pathway, allowing for repeated, deep interactions between text tokens $\mathbf{C}$ and occupancy tokens $\mathbf{H}$ to strengthen fine-grained conditioning.

To efficiently process the massive number of occupancy tokens, we factorize the attention mechanism via Spatio-Temporal Separated Attention (STSA) \cite{STSA-Latte__ma2024latte}. Specifically, each transformer block decomposes the operation into two sequential steps:
(i) \emph{Frame-wise 3D spatial joint attention}, which models interactions between text tokens and spatial occupancy tokens within each independent frame to ensure script-to-layout grounding; followed by
(ii) \emph{Patch-wise temporal self-attention}, applied solely across the temporal axis of occupancy tokens to propagate dynamics smoothly over time.
To align with this factorization, we apply 2D spatial RoPE \cite{rope__su2024roformer} on the $H{\times}W$ grid within each frame (with the $Z$-axis folded into the channel dimension), and 1D temporal RoPE across frames.

\subsection{Interaction-Consistent Forecasting via History Anchoring}\label{sec:objective}
To train \ModelName{} for stable and high-quality continuous latent generation, we adopt the Continuous Normalizing Flow (CNF) framework via Flow Matching \cite{FlowMatching__lipman2022flow}. 
We define $t=0$ as the ground-truth clean data and $t=1$ as the pure noise. 
Let $\mathbf{x}_0$ denote the ground-truth latent occupancy tokens and $\mathbf{x}_1 \sim \mathcal{N}(\mathbf{0}, \mathbf{I})$ be the standard Gaussian noise. We construct a linear probability path $\mathbf{x}_t = (1-t)\mathbf{x}_0 + t\mathbf{x}_1$, where $t \sim \mathcal{U}(0,1)$. The corresponding vector field is given by $v_t(\mathbf{x}) = \mathbf{x}_1 - \mathbf{x}_0$. 
The base objective trains the model $v_\theta$ to regress this vector field conditioned on the refined text tokens $\mathbf{C}$.

\textbf{Training with History-Prefix Anchoring.} For behavior generation, users often specify partial layouts or history contexts $\mathbf{O}_{1:h}$. Maintaining strict consistency with these constraints is critical, as small deviations in early frames can compound into collisions or interaction drift. 
Rather than relying on training-free heuristics during inference (which often yield temporal inconsistencies), we explicitly equip the model with forecasting capabilities during the training phase.

To achieve this, we introduce a stochastic \emph{token-replace anchoring} strategy, visualized in \cref{fig:model_architecture}(c) \cite{hunyuanvideo__kong2024hunyuanvideo}. During training, we randomly sample a history horizon $h$. With a certain probability (similar to classifier-free guidance dropping), we replace the noisy history prefix $\mathbf{x}_t^{(1:h)}$ with the perfectly clean ground-truth tokens $\mathbf{x}_0^{(1:h)}$. 
Let $\tilde{\mathbf{x}}_t$ denote this partially clean, partially noisy input sequence. To ensure the model focuses entirely on forecasting the unknown dynamics, we apply a masked loss that only penalizes the predictions on the future frames $(h+1:F)$:
\begin{equation}
    \mathcal{L}_{\mathrm{Anchor}} = \mathbb{E}_{t, \mathbf{x}_0, \mathbf{x}_1} \left[ \left\| v_\theta(\tilde{\mathbf{x}}_t, \mathbf{C}, t)^{(h+1:F)} - (\mathbf{x}_1 - \mathbf{x}_0)^{(h+1:F)} \right\|_2^2 \right].
    \label{eq:anchoring_loss}
\end{equation}
This formulation forces the network to learn the conditional transition dynamics from a clean history to a noisy future, strictly guided by the script $\mathbf{C}$.

\textbf{Inference via Step-wise Anchoring.} During the inference (ODE sampling) phase, we start from pure noise $\mathbf{x}_1$ and integrate backwards to $t=0$. To enforce robust controllability and ensure the generated sequence perfectly matches the user-provided history $\mathbf{H}_{1:h}$, we apply step-wise anchoring. 
At every integration step $t$, after the solver updates the sequence, we forcefully overwrite the history prefix with the clean condition $\mathbf{x}_{t-\Delta t}^{(1:h)} \leftarrow \mathbf{H}_{1:h}$. 
Because the model has been explicitly trained to condition on clean history prefixes via \cref{eq:anchoring_loss}, this step-wise replacement seamlessly anchors the generation process without introducing out-of-distribution artifacts, ensuring that the future dynamics $\mathbf{x}_0^{(h+1:F)}$ remain strictly coherent with the provided context.

\pdfbookmark[1]{4 OccInteract-85k: A Language-Conditioned Occupancy Dataset}{main.sec.dataset}
\section{\DatasetName{}: A Language-Conditioned Occupancy Dataset}
\label{sec:data_pipeline}

To train \ModelName{} with language-conditioned occupancy prediction, we require paired supervision between (i) 4D occupancy representations and (ii) natural-language descriptions that emphasize \emph{geometry, topology, and interaction intent} rather than appearance.
To this end, we introduce \textbf{\DatasetName{}}, a dataset constructed via an automated pipeline that synthesizes three complementary corpora: \emph{static layout captions} (\StaticDatasetName{}), \emph{environment-interaction clips} (\EnvDatasetName{}), and \emph{agent-interaction simulations} (\AgentDatasetName{}).
All annotations are generated with structured prompting and enforced output schemas, enabling scalable production and automatic filtering.

\subsection{Data Curation Pipeline}
\label{sec:data_sources}
Our pipeline integrates three data sources to balance realism, scale, and diversity:
(i) \emph{Real-world Logs.} We leverage UniOcc \cite{uniocc__wang2025uniocc} to aggregate real-world autonomous driving logs, providing unified occupancy representations with high photorealism.
(ii) \emph{Large-scale Simulation.} To overcome the scalability limits of real logs, we collect extensive simulator-based data following the CarlaSC protocol \cite{CarlaSC__wilson2022motionsc}, enabling diverse environmental configurations.
(iii) \emph{Safety-critical Scenarios.} To target long-tail and safety-critical interactions, which are often absent in real logs. Therefore we generate text-defined scenarios using TTSG \cite{ttsg__ruan2024traffic}. These are rendered via CarlaSC, ensuring coverage of high-risk edge cases.

\paragraph{Standardization.}
We standardize these raw sources into paired \emph{occupancy--language} training instances. Following UniOcc \cite{uniocc__wang2025uniocc}, we standardize all data sources into a unified occupancy grid format. Then, we render semantic occupancy grids into BEV sequences and prompt a frozen VLM (\texttt{qwen3-vl-235b-a22b-instruct}) to generate schema-constrained, machine-parseable JSON captions. This process yields annotations at three temporal granularities: (i) single-frame layout captions for learning geometric priors, (ii) clips around key maneuvers for learning behavior-conditioned structure, and (iii) specified multi-agent plans for generating diverse interactions, forming the sub-datasets detailed below. Table~\ref{tab:processed_dataset_stats} summarizes the statistics. Concretely, all sources are unified to a $200{\times}200{\times}16$ voxel grid ($0.4\,\mathrm{m/voxel}$, $X/Y\colon{\pm}40\,\mathrm{m}$, $Z\colon{-3.2}$--$3.2\,\mathrm{m}$) sampled at $2\,\mathrm{Hz}$, with semantic labels remapped to a shared 11-category ontology following UniOcc~\cite{uniocc__wang2025uniocc}. 

\begin{table}[t]
\centering
\small
\setlength{\tabcolsep}{6pt}
\caption{\textbf{Dataset Statistics.} Overview of the processed corpora. Instances denote single frames for \StaticDatasetName{} and video clips for \EnvDatasetName{} / \AgentDatasetName{}.}
\label{tab:processed_dataset_stats}
\vspace{\tabcapgap}
\begin{tabular}{@{}lrr@{}}
\toprule
Dataset & \# Instances & Total Duration \\
\midrule
\StaticDatasetName{} & 73.5k & N/A (Single-frame) \\
\EnvDatasetName{} & 7.6k & 128\,h \\
\AgentDatasetName{} & 4.6k & 12\,h \\
\bottomrule
\end{tabular}
\vspace{\tabbottomgap}
\end{table}

\subsection{Dataset Composition}

\paragraph{\StaticDatasetName{}: Static Layout Priors.} 
This subset pairs single-frame BEV semantic occupancy snapshots with ego-centric layout captions. The annotations focus on road topology (e.g., lane structures, intersections) and drivable regions, providing fundamental geometric priors independent of temporal dynamics.

\paragraph{\EnvDatasetName{}: Behavior-Conditioned Dynamics.} 
Targeting key maneuvers at junctions, this subset provides clip-level supervision. Each instance includes a discrete ego-action command (straight/left/right) and a description of the local topology. This facilitates learning how occupancy structures evolve conditioned on ego-behavior.

\begin{figure}[t] 
    \centering
    \includegraphics[width=\linewidth]{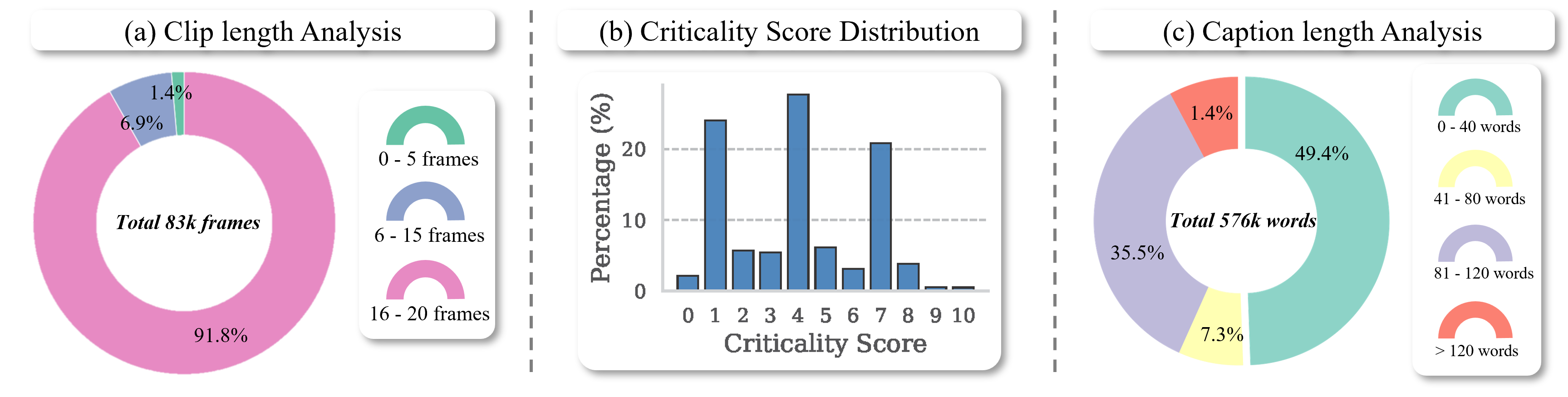}
    \vspace{\figcapgap}
    \caption{\textbf{Data Statistics of \AgentDatasetName{}.} (a) Distribution of clip lengths. (b) Histogram of VLM-predicted criticality scores. (c) Distribution of caption lengths.} 
    \label{fig:criticality_score_distribution} 
    \label{fig:AgentDataset_data_distribution}
    \vspace{\figbottomgap}
\end{figure}

\paragraph{\AgentDatasetName{}: Multi-Agent Interactions.} 
This corpus covers a spectrum from routine driving to safety-critical near-misses. Each instance is paired with an interaction-focused caption and a \emph{criticality score} ($s \in [0,10]$), predicted by a frozen VLM judge (\texttt{qwen3-vl-235b-a22b-instruct}). 
We interpret this score as a proxy for risk: a low score corresponds to routine, non-conflicting interactions; medium indicates potential conflicts requiring caution (e.g., nearby vulnerable road users or occlusions); high reflects imminent hazards or near-collision situations.
As shown in \cref{fig:AgentDataset_data_distribution} (b), the scores exhibit a trimodal distribution (peaks at 1/4/7), effectively stratifying samples into low, medium, and high-risk regimes. This metadata enables curriculum learning and targeted sampling of rare safety-critical scenarios.
Furthermore, \cref{fig:AgentDataset_data_distribution} (a) and (c) present the statistics for clip durations and caption lengths.

\subsection{Automated Quality Assessment} 
\label{sec:automated_quality_assessment}
\label{sec:quality_scoring}

To close the loop between data generation and model learning, we implement a rubric-based VLM evaluator. This mechanism serves a dual purpose: filtering low-quality synthetic data and benchmarking model performance (see \cref{sec:experiments}).

\paragraph{Evaluation Protocol.}
We adopt a reference-free evaluation paradigm. Specifically, we render the generated semantic occupancy grids into BEV video clips and prompt a frozen VLM judge to score them against the source text descriptions. 
We define three evaluation axes, scored on an integer scale of $[1,5]$:
(i) \emph{scene completeness} (coverage of relevant static/dynamic elements), (ii) \emph{structural correctness} (physical plausibility and topological consistency), and (iii) \emph{Semantic Alignment} (faithfulness to the prompt).

To ensure robustness, we employ three state-of-the-art VLMs as judges: \texttt{gemini-3.0-pro}, \texttt{gpt-5.2}, and \texttt{qwen3-vl-235b-a22b-instruct}. 
We sample 100 instances from each sub-dataset, with each instance evaluated five times to reduce variance.

\begin{table}[t]
\centering
\small
\setlength{\tabcolsep}{5pt}
\caption{\textbf{Automated Data Validation.} Mean rubric scores ($1\text{-}5$) assessing the quality of the generated dataset.}
\label{tab:quality_scoring_results}
\vspace{\tabcapgap}
\resizebox{\linewidth}{!}{%
\begin{tabular}{@{}llcccc@{}}
\toprule
Dataset & Judge VLM & Completeness $\uparrow$ & Structural $\uparrow$ & Sem. Align. $\uparrow$ & Mean $\uparrow$ \\
\midrule
\multirow{3}{*}{\StaticDatasetName{}} & \texttt{gemini-3.0-pro} & 4.16 & 4.20 & 3.84 & 4.08 \\
 & \texttt{gpt-5.2} & 2.97 & 3.71 & 3.03 & 3.24 \\
 & \texttt{qwen3-vl-235b-a22b-instruct} & 3.22 & 3.84 & 3.97 & 3.68 \\
\midrule
\multirow{3}{*}{\EnvDatasetName{}} & \texttt{gemini-3.0-pro} & 4.62 & 4.60 & 3.55 & 4.26 \\
 & \texttt{gpt-5.2} & 3.21 & 3.78 & 3.01 & 3.33 \\
 & \texttt{qwen3-vl-235b-a22b-instruct} & 3.32 & 3.81 & 3.66 & 3.60 \\
\midrule
\multirow{3}{*}{\AgentDatasetName{}} & \texttt{gemini-3.0-pro} & 4.51 & 4.53 & 3.49 & 4.18 \\
 & \texttt{gpt-5.2} & 3.21 & 3.72 & 2.58 & 3.17 \\
 & \texttt{qwen3-vl-235b-a22b-instruct} & 3.30 & 3.64 & 3.32 & 3.42 \\
\bottomrule
\end{tabular}%
}
\vspace{\tabbottomgap}
\end{table}

\paragraph{Analysis.}
Table~\ref{tab:quality_scoring_results} presents a comprehensive evaluation across three axes. 
First, regarding \textit{absolute quality}, \texttt{gemini-3.0-pro} consistently awards high scores (mean $>4.0$), confirming that our synthesized occupancy scenes possess high completeness and physical plausibility. 
Second, regarding \textit{scorer calibration}, we observe a systematic offset between judges; for instance, \texttt{gpt-5.2} is notably stricter (mean $\approx 3.2$). We attribute this to the inherent visual abstraction of voxelized representations, which stricter VLMs may penalize compared to photorealistic imagery. 
Third, and most importantly, we demonstrate \textit{pipeline robustness} through relative consistency. Despite the increasing complexity from static layouts (\StaticDatasetName{}) to multi-agent interactions (\AgentDatasetName{}), the score variance under the same judge remains minimal (e.g., \texttt{gpt-5.2} means range tightly: $3.17\text{--}3.33$). 
This stability indicates that our automated pipeline maintains uniform fidelity regardless of scenario complexity, validating the reliability of \DatasetName{} for large-scale training.

\pdfbookmark[1]{5 Experiments}{main.sec.experiments}
\section{Experiments}
\label{sec:experiments}

\subsection{Experimental Setup}
\paragraph{Data.}
We evaluate on the same processed dataset described in \cref{sec:data_pipeline}, which contains about 85k training instances.

\paragraph{Compared methods.}   
We compare \ModelName{} with recent state-of-the-art 4D occupancy generation methods, including DOME \cite{dome__gu2024dome}, COME \cite{come__shi2025come}, and DynamicCity \cite{dynamiccity__bian2024dynamiccity}. As discussed in \cref{sec:intro}, these traditional approaches rely on explicit geometric conditions, such as predefined ego-vehicle trajectories or semantic layouts, to control the generation process. 
To the best of our knowledge, \ModelName{} is the pioneering framework for purely text-guided 4D occupancy generation. Lacking direct off-the-shelf baselines, we adapt DynamicCity by replacing its geometric control tokens with text embeddings from pre-trained encoders, establishing two baselines: \textit{DynamicCity + CLIP \cite{clip_radford2021learning}} and \textit{DynamicCity + T5 \cite{t5__raffel2020exploring}}.
Finally, we report results for two variants of our proposed model: (i) w/o history-prefix control, and (ii) w/ history-prefix control.   

\paragraph{Implementation Details.}
We implement \ModelName{} using PyTorch, trained for $100\mathrm{k}$ iterations with batch size 32, AdamW optimizer ($\mathrm{lr}{=}1{\times}10^{-4}$, $\beta_1{=}0.9$, $\beta_2{=}0.95$), and CFG dropout probability 0.15. The MMDiT backbone has 14 transformer layers ($d_{\mathrm{model}}{=}896$) with a 2-layer Token Refiner. At inference we use 20 Euler steps with CFG scale 7.5.

\paragraph{Evaluation Metrics.}
Following previous work \cite{semcity__lee2024semcity,dynamiccity__bian2024dynamiccity,X-scene__yang2025x}, we render the 3D/4D occupancy grids into BEV-view RGB images and clips. These rendered RGB representations are subsequently utilized to evaluate all single-frame and video-level visual quality metrics.
Specifically, we report Inception Score (IS) \cite{InceptionScore__barratt2018note}, Fr\'echet Inception Distance (FID) \cite{fid__heusel2017gans}, Kernel Inception Distance (KID) \cite{kid_binkowski2018demystifying} and Fr\'echet Video Distance (FVD) \cite{fvd_unterthiner2018towards} to assess fidelity and temporal coherence, alongside Precision and Recall for diversity.
For instruction alignment, we employ the VLM-based instruction alignment score (\cref{sec:quality_scoring}). To ensure consistent and rigorous comparison, we fix \texttt{gemini-3.0-pro} as our sole judge VLM for all experiments.
For evaluation, we uniformly sample $10\mathrm{K}$ ground-truth 16-frame clips and generate a matching set via the CFG unconditional branch. Single-frame metrics (IS, FID, KID, Precision, Recall) are computed on 5 frames uniformly sampled per clip using Inception-v3~\cite{Inception-v3__szegedy2016rethinking} features; video-level FVD and video Precision/Recall use the full 16-frame sequences with I3D~\cite{i3d__carreira2017quo} features.

\subsection{Generation Quality Results}

\begin{table}[t]
\centering
\small
\setlength{\tabcolsep}{5pt}
\caption{\textbf{Single-frame visual quality.} We evaluate per-frame image fidelity and diversity using Inception Score (IS), Fr\'echet Inception Distance (FID), Kernel Inception Distance (KID), Precision and Recall.}
\label{tab:main_results_frame_quality}
\vspace{\tabcapgap}
\resizebox{\linewidth}{!}{%
\begin{tabular}{@{}lccccc@{}}
\toprule
Method & IS $\uparrow$ & FID $\downarrow$ & KID $\downarrow$ & Precision $\uparrow$ & Recall $\uparrow$ \\   
\midrule  
DOME \cite{dome__gu2024dome} & 3.21 & 67 & 0.06 & 0.38 & \textbf{0.23} \\   
COME \cite{come__shi2025come} & 3.26 & 77 & 0.06 & \textbf{0.40} & \textbf{0.23} \\   
DynamicCity \cite{dynamiccity__bian2024dynamiccity} & \textbf{3.39} & \textbf{61} & \textbf{0.05} & 0.31 & \textbf{0.23} \\   
\midrule  
DynamicCity + CLIP & 3.39 & 65 & \textbf{0.05} & 0.32 & \textbf{0.24} \\   
DynamicCity + T5 & 3.36 & 66 & \textbf{0.05} & 0.31 & 0.22 \\   
\ModelName{} w/o history-prefix control & 3.39 & 64 & \textbf{0.05} & \textbf{0.34} & \textbf{0.24} \\   
\ModelName{} w/ history-prefix control & \textbf{3.40} & \textbf{62} & \textbf{0.05} & 0.33 & 0.23 \\   
\bottomrule  
\end{tabular}   
}%
\vspace{\tabbottomgap}
\end{table}   

\paragraph{Quantitative results.}
We present the single-frame visual quality results in \cref{tab:main_results_frame_quality} and the video generation quality results in \cref{tab:main_results_quality}.   

As shown in \cref{tab:main_results_frame_quality}, \ModelName{} demonstrates highly competitive per-frame generation quality. 
It is important to note that traditional methods (e.g., DOME, COME, DynamicCity) rely on explicit geometric conditions (e.g., trajectories or layouts) which naturally constrain the generation space and aid in structural consistency. 
Despite \ModelName{} being designed for the more abstract text-to-4D task, our unconditional samples achieve an Inception Score (IS) of 3.40 and FID of 62, performance that is comparable to these geometric-guided upper bounds. 
Crucially, when compared to the direct text-adapted baselines (DynamicCity + CLIP/T5) which share the same functional goal, our method consistently achieves superior performance across all single-frame metrics. 
This confirms that our architecture effectively learns the underlying scene distribution and generates high-fidelity occupancy frames, outperforming naive adaptations of previous works.

\begin{table}[t]
\centering
\small
\setlength{\tabcolsep}{6pt}
\caption{\textbf{Video generation quality and instruction alignment.} We evaluate video-level fidelity and diversity using Fr\'echet Video Distance (FVD), Precision and Recall. Instruction alignment is evaluated using the VLM-based rubric score (\cref{sec:quality_scoring}).}
\label{tab:main_results_quality}
\vspace{\tabcapgap}
\resizebox{\linewidth}{!}{%
\begin{tabular}{@{}lccc|cccc@{}}
\toprule  
& \multicolumn{3}{c|}{Video realism} & \multicolumn{4}{c}{Instruction alignment} \\   
\cmidrule(lr){2-4} \cmidrule(lr){5-8}   
Method & FVD $\downarrow$ & Precision $\uparrow$ & Recall $\uparrow$ & Completeness $\uparrow$ & Structural $\uparrow$ & Sem. Align. $\uparrow$ & Mean $\uparrow$ \\   
\midrule  
DOME \cite{dome__gu2024dome} & 437 & 0.55 & 0.03 & N/A & N/A & N/A & N/A \\   
COME \cite{come__shi2025come} & \textbf{391} & 0.56 & 0.04 & N/A & N/A & N/A & N/A \\   
DynamicCity \cite{dynamiccity__bian2024dynamiccity} & 401 & \textbf{0.57} & \textbf{0.08} & N/A & N/A & N/A & N/A \\   
\midrule  
DynamicCity + CLIP & 388 & 0.60 & 0.008 & 3.03 & 3.54 & 2.69 & 3.09 \\   
DynamicCity + T5 & 396 & \textbf{0.64} & 0.02 & 2.84 & 3.34 & 2.73 & 2.97 \\
\ModelName{} w/o history-prefix control & 354 & 0.48 & \textbf{0.12} & 3.59 & 3.66 & 3.04 & 3.43 \\   
\ModelName{} w/ history-prefix control & \textbf{275} & 0.58 & \textbf{0.12} & \textbf{3.81} & \textbf{4.01} & \textbf{3.32} & \textbf{3.73} \\   
\bottomrule  
\end{tabular}   
}%
\vspace{\tabbottomgap}
\end{table}

For video-level generation (\cref{tab:main_results_quality}, left), \ModelName{} significantly improves temporal coherence and dynamic realism. 
Our full model achieves an FVD of 275, outperforming the strongest baseline (DynamicCity + CLIP, FVD 388) by a remarkable 29.1\%. 
Furthermore, \ModelName{} achieves the highest video Recall (0.12) while maintaining a strong Precision (0.58). 
We observe an interesting phenomenon with the text-adapted baselines (DynamicCity + CLIP/T5): while they achieve relatively high Precision (e.g., 0.64 for T5), their Recall is extremely low (e.g., 0.008 for CLIP).
This suggests a severe mode collapse, where standard text encoders fail to adequately capture complex text instructions, leading the model to repeatedly generate a limited set of ``safe'' or static scenarios.
In contrast, our method generates spatio-temporal occupancy sequences that are not only realistic but also encompass a significantly wider diversity of dynamic scenarios.

Furthermore, in terms of text-to-occupancy instruction alignment (\cref{tab:main_results_quality}, right), \ModelName{} substantially surpasses the text-adapted baselines across all VLM-based rubric scores. For instance, our full model achieves a mean score of 3.73. The low semantic alignment scores of DynamicCity + CLIP/T5 further corroborate our observation regarding their low Recall: standard text encoders struggle to differentiate between diverse prompts, resulting in poor alignment and limited generation diversity. This highlights the severe semantic-spatiotemporal gap in standard text encoders and proves the superiority of our VLM-driven architecture. Notably, the inclusion of the history-prefix control mechanism further boosts the structural correctness and instruction-following capabilities (improving the mean score from 3.43 to 3.73), demonstrating the effectiveness of our proposed conditioning design for complex multi-agent scenario orchestration.

\begin{figure}[t]
    \centering
    \includegraphics[width=\linewidth]{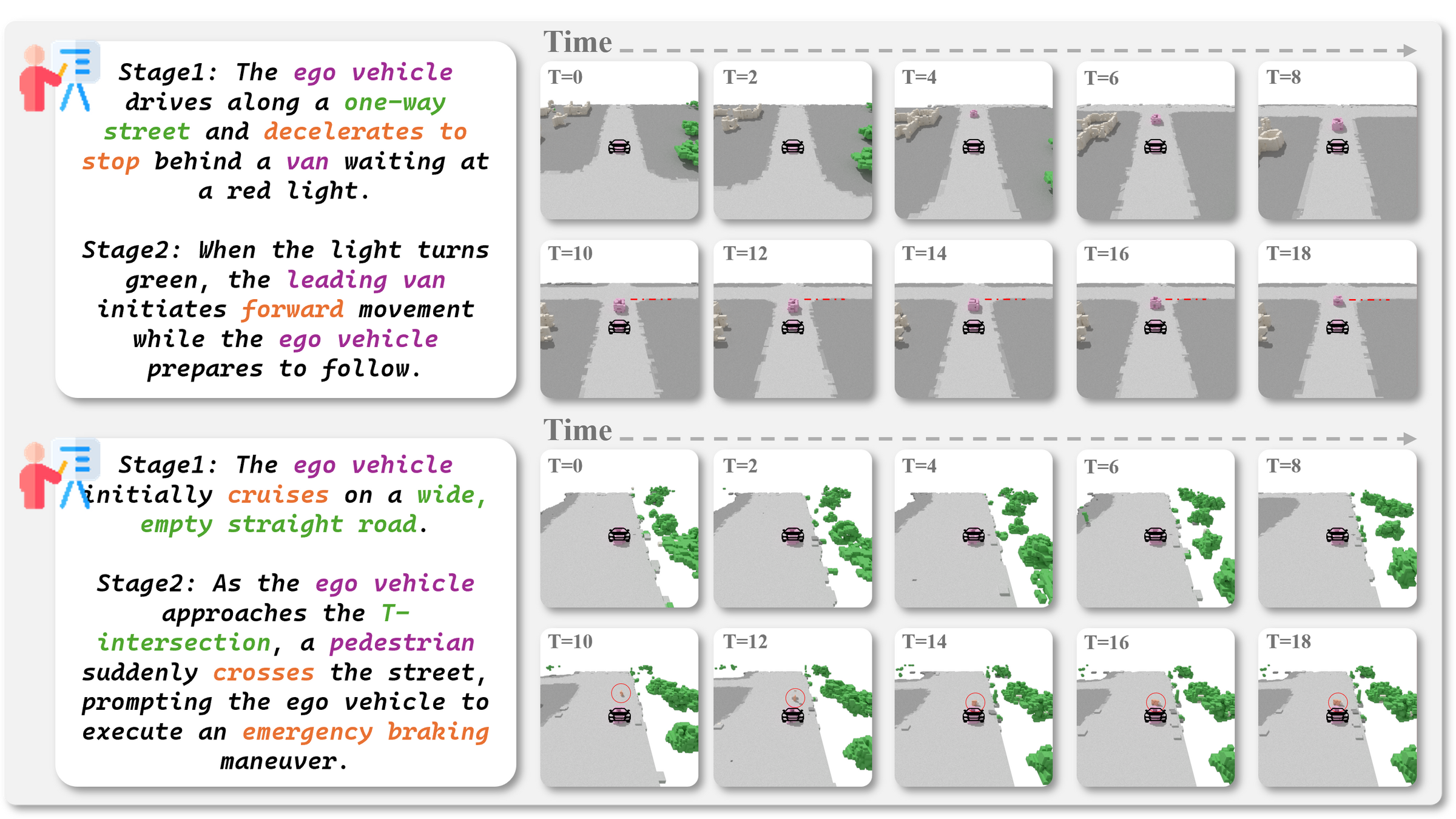}
    \vspace{\figcapgap}
    \caption{\textbf{Long-horizon narrative generation.} We visualize multi-stage rollouts for a stop-and-go scenario (top) and an emergency braking event (bottom).}
    \label{fig:qual_history}
    \vspace{\figbottomgap}
\end{figure}

\paragraph{Qualitative results on history-prefix control.}
As the teaser in \cref{sec:intro} highlights \ModelName{}'s diverse controllability, we further investigate its capability in long-horizon autoregressive rollouts. \cref{fig:qual_history} illustrates two multi-stage narratives where the ego vehicle is denoted by a \textit{black car icon}. 
In the \textbf{stop-and-go scenario} (top), the model follows the instruction to decelerate on a one-way street and stop behind a van at a red light (Stage 1). In Stage 2, responding to the ``green light'' prompt, the van initiates movement across the intersection edge (marked by \textit{red dashed lines}), correctly prompting the ego to prepare to follow.
In the \textbf{emergency braking scenario} (bottom), the ego initially cruises on a wide, empty straight road. Subsequently, the model successfully generates a sudden pedestrian crossing event at a T-intersection (highlighted by a \textit{red circle}), forcing the ego to execute an emergency stop.
These results confirm that history-prefix control effectively maintains spatio-temporal coherence and causal logic across generation stages.

\begin{figure}[t]
    \centering
    \includegraphics[width=\linewidth]{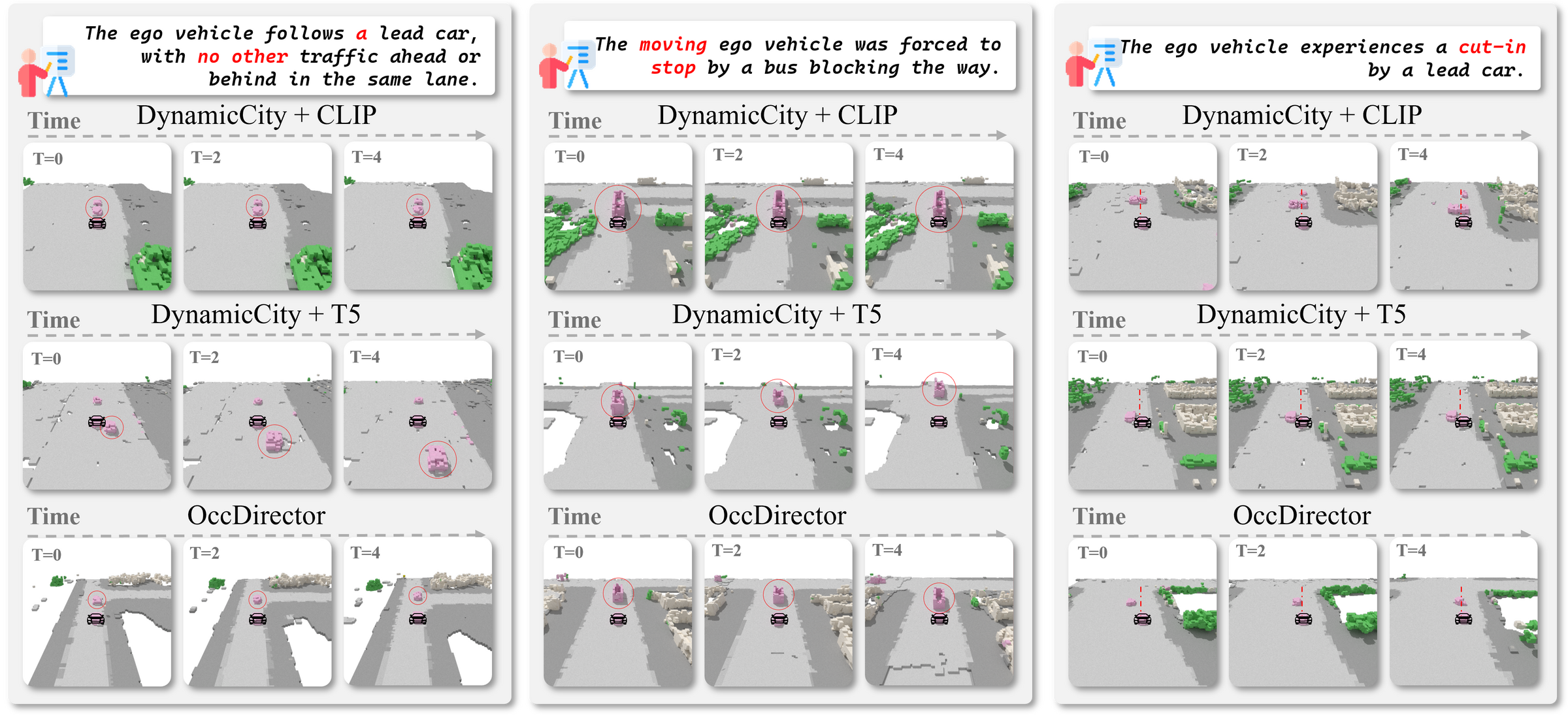}
    \vspace{\figcapgap}
    \caption{\textbf{Qualitative comparison of text-to-occupancy alignment.} We evaluate adherence to fine-grained instructions: (Left) Negation and cardinality; (Middle) Dynamic attribute binding; (Right) Complex ``cut-in'' trajectory.}
    \label{fig:qual_comparison}
    \vspace{\figbottomgap}
\end{figure}

\paragraph{Qualitative comparison on fine-grained text alignment.}
\cref{fig:qual_comparison} compares \ModelName{} with text-adapted baselines on three prompts requiring precise semantic understanding.
In the first column ``...follows a lead car, with \textit{no other} traffic...'', baselines fail to respect the negation, generating extraneous vehicles. We highlight these hallucinations in baselines versus the correct single lead car in our result using \textit{red circles}.
In the second column ``...moving ego... \textit{stop} by a bus'', baselines exhibit attribute confusion (e.g., DynamicCity + CLIP renders a completely static scene, while the T5 variant incorrectly halts the ego vehicle while moving the bus). Here, \textit{red circles} contrast the baselines' incorrect objects with our successfully generated blocking bus.
Finally, for the ``cut-in'' prompt, we indicate the ego's forward direction with \textit{red dashed lines}. While baselines revert to simple lane-following, only \ModelName{} synthesizes the lead car's lateral intrusion across this line, demonstrating superior grasp of spatial dynamics.

\subsection{Ablation Study} 
\label{sec:ablation}

To thoroughly evaluate the effectiveness of our proposed architectural designs, we conduct comprehensive ablation studies. Since the impact of the history-prefix control has already been demonstrated in \cref{tab:main_results_quality} and \cref{fig:qual_history}, we establish our ablation baseline using the text-to-occupancy generation setting \textbf{without (w/o) history control}. 
We focus on two key components: 
(i) The backbone strategy, where we compare our Spatio-Temporal Separated Attention (STSA) against a Full Attention baseline that applies standard joint attention over the entire flattened 4D token sequence. 
(ii) The Token Refiner, a bidirectional transformer module designed to bridge the semantic-spatial gap of VLM features. 
Results are summarized in \cref{tab:ablation_architecture}. 

\begin{table}[t]
\centering
\small
\setlength{\tabcolsep}{5pt}
\caption{\textbf{Ablation study on model architecture.} We evaluate the impact of the attention mechanism (Full vs. STSA) and the Token Refiner under the setting w/o history control.}
\label{tab:ablation_architecture}
\vspace{\tabcapgap}
\resizebox{\linewidth}{!}{%
\begin{tabular}{@{}lc|ccc|cccc@{}}
\toprule
\multicolumn{2}{c|}{Architecture Design} & \multicolumn{3}{c|}{Video realism} & \multicolumn{4}{c}{VLM score $\uparrow$} \\ 
\cmidrule(lr){1-2} \cmidrule(lr){3-5} \cmidrule(lr){6-9} 
Backbone & Token Refiner & FVD $\downarrow$ & Precision $\uparrow$ & Recall $\uparrow$ & Completeness $\uparrow$ & Structural $\uparrow$ & Instr.~follow $\uparrow$ & Mean $\uparrow$ \\ 
\midrule
Full attention & $\times$ & 592 & \textbf{0.48} & 0.02 & \textbf{2.98} & \textbf{2.86} & 2.48 & 2.77 \\ 
Full attention & \checkmark & \textbf{532} & 0.41 & \textbf{0.07} & 2.97 & 2.74 & \textbf{2.63} & \textbf{2.78} \\ 
\midrule
STSA (ours) & $\times$ & 407 & 0.49 & \textbf{0.14} & 3.46 & 3.39 & 2.86 & 3.24 \\ 
STSA (ours) & \checkmark & \textbf{354} & \textbf{0.58} & 0.12 & \textbf{3.59} & \textbf{3.66} & \textbf{3.04} & \textbf{3.43} \\ 
\bottomrule
\end{tabular} 
}%
\vspace{\tabbottomgap}
\end{table}

\paragraph{Effectiveness of STSA and Token Refiner.} 
First, STSA is critical for modeling 4D dynamics. Compared to the Full Attention baseline (FVD 532), STSA significantly reduces FVD to 354 and improves the VLM score (2.78 $\to$ 3.43), confirming the benefit of factorizing spatial and temporal computation.
Second, the Token Refiner effectively bridges the gap between frozen text embeddings and occupancy features. Removing the refiner degrades performance (FVD $354 \to 407$), verifying its necessity for aligning high-dimensional text features with spatial constraints.
Notably, the Full Attention backbone fails to fully leverage the refined tokens (VLM score stagnates at 2.78), whereas our STSA architecture successfully integrates them to achieve state-of-the-art fidelity and alignment.

\pdfbookmark[1]{6 Conclusion}{main.sec.conclusion}
\section{Conclusion}

In this paper, we presented \ModelName{}, a pioneering framework capable of generating dynamic 4D occupancy scenes purely from natural language instructions. By shifting the paradigm from rigid geometric conditioning to high-level semantic orchestration, \ModelName{} acts as a ``scenario director'', enabling intuitive control over complex multi-agent behaviors and environmental interactions. To achieve this, we introduced a VLM-driven Spatio-Temporal MMDiT architecture equipped with a novel history-prefix anchoring strategy, effectively bridging the semantic-spatiotemporal gap and ensuring physically plausible, interaction-consistent rollouts. Furthermore, we contributed \DatasetName{}, a large-scale dataset with multi-level language descriptions, alongside a rigorous VLM-based evaluation benchmark. Extensive experiments demonstrate that our approach achieves state-of-the-art generation quality and unprecedented instruction-following capabilities.

\newpage
\bibliographystyle{splncs04}
\bibliography{main}

\let\addcontentsline\oldaddcontentsline


\title{\ModelName{}: Language-Guided Behavior and Interaction Generation in 4D Occupancy Space}
\subtitle{Supplementary Material}

\titlerunning{Supplementary Material}
\authorrunning{Z.~Liang, T.~Yan et al.}

\author{Zhuding Liang\inst{1}\thanks{Equal Contribution.} \and
Tianyi Yan\inst{1}\textsuperscript{*} \and
Dubing Chen\inst{1} \and
Jiasen Zheng\inst{2} \and
Huan Zheng\inst{1} \and
Cheng-zhong Xu\inst{1} \and
Yida Wang\inst{3} \and
Kun Zhan\inst{3} \and
Jianbing Shen\inst{1}\thanks{Corresponding Author.}}

\institute{SKL-IOTSC, CIS, University of Macau, Macau, China \and Northwestern University, Evanston, United States\and
Li Auto Inc, Beijing, China}

\let\oldaddcontentsline\addcontentsline
\renewcommand{\addcontentsline}[3]{}
\maketitle
\let\addcontentsline\oldaddcontentsline

\pdfbookmark[0]{Supplementary Material}{supp.title}

\addtocontents{toc}{\protect\setcounter{tocdepth}{2}}

\begin{center}
    {\large \bfseries Table of Contents}
\end{center}
\hrule
\vspace{-24pt}
\begingroup
\renewcommand{\clearpage}{}
\renewcommand{\contentsname}{}
\tableofcontents
\endgroup
\vspace{16pt}
\hrule
\clearpage

\appendix
\setcounter{figure}{0}
\setcounter{table}{0}
\setcounter{equation}{0}
\renewcommand{\thefigure}{S\arabic{figure}}
\renewcommand{\thetable}{S\arabic{table}}
\renewcommand{\theequation}{S\arabic{equation}}



\section{Additional Implementation Details}
\label{sec:supp_impl}

\subsection{\ModelName{} Components and Hyperparameters}
\label{sec:supp_impl_hparams}

\paragraph{1. Joint Latent Representation.}
We formalize the derivation of the occupancy tokens $\mathbf{H}$ and the text tokens $\mathbf{C}$.

\textbf{Occupancy Tokenization:} Given a discrete 4D occupancy rollout $\mathbf{O}_{1:F} \in \{0,\dots,K-1\}^{F \times 200 \times 200 \times 16}$, we first linearly interpolate the spatial dimensions to $F \times 224 \times 224 \times 16$ to ensure compatibility with standard $2\times 2$ patchification. The frozen Occ-VAE maps each occupancy voxel ID to a dense embedding via a pre-trained lookup table $\mathrm{Emb}(\cdot)$ with expansion ratio $e=8$:
\begin{equation}
    \mathbf{E} = \mathrm{Emb}(\mathbf{O}_{1:F}) \in \mathbb{R}^{F \times 224 \times 224 \times 16 \times 8}.
\end{equation}
To achieve the efficient ``3D spatial tokens'' representation, we fold the depth/height bins $D=16$ into the channel axis, obtaining a channel-first latent $\mathbf{X} \in \mathbb{R}^{128 \times F \times 224 \times 224}$. This high-dimensional spatial representation is then compressed by the Occ-VAE encoder to a continuous spatio-temporal latent grid $\mathbf{z} \in \mathbb{R}^{16 \times F \times 28 \times 28}$ via reparameterization:
\begin{equation}
    \mathbf{z} = \boldsymbol\mu + \boldsymbol\sigma \odot \boldsymbol\epsilon, \quad \boldsymbol\epsilon \sim \mathcal{N}(\mathbf{0},\mathbf{I}).
\end{equation}
To interface with the transformer, we apply a spatial patchification (Unfold) with a patch size $p=2$, resulting in $S = (28/2)^2 = 196$ spatial tokens per frame. These tokens are projected to the model dimension $d_{\mathrm{model}}=896$ via a linear layer.

\textbf{Text Token Refinement:} We utilize Qwen3-VL-2B \cite{supp_qwen3__yang2025qwen3} as \ModelName{}'s frozen VLM. It extracts raw text features $\tilde{\mathbf{C}} \in \mathbb{R}^{L \times d_{\mathrm{text}}}$ from the prompt $\pi$. The Token Refiner $\mathrm{Refine}(\cdot)$ is a lightweight bidirectional transformer consisting of $N_{\mathrm{ref}}$ blocks. Let $\mathbf{C}^{(0)} = \tilde{\mathbf{C}} \mathbf{W}_{\mathrm{in}}$ be the initial projection to $d_{\mathrm{model}}$. Each block $i \in \{1, \dots, N_{\mathrm{ref}}\}$ performs: 
\begin{align} 
    \hat{\mathbf{C}}^{(i)} &= \mathbf{C}^{(i-1)} + \mathrm{MHSA}(\mathrm{LN}(\mathbf{C}^{(i-1)})), \\
    \mathbf{C}^{(i)} &= \hat{\mathbf{C}}^{(i)} + \mathrm{MLP}(\mathrm{LN}(\hat{\mathbf{C}}^{(i)})),
\end{align}
where $\mathrm{MHSA}$ is an unmasked, bidirectional multi-head self-attention to capture global semantic relations. The final refined text tokens are $\mathbf{C} = \mathbf{C}^{(N_{\mathrm{ref}})} \in \mathbb{R}^{L \times d_{\mathrm{model}}}$.

\paragraph{2. Scalable 4D Generation Backbone.} 
We detail the internal structure of the $l$-th transformer block. The input consists of the packed occupancy tokens $\mathbf{H}^{(l-1)} \in \mathbb{R}^{F \times S \times d_{\mathrm{model}}}$ and the text tokens $\mathbf{C}^{(l-1)} \in \mathbb{R}^{L \times d_{\mathrm{model}}}$. The data flow proceeds through four ordered stages. 

\textbf{(I) Pre-Modulation.} 
Given the integration timestep $t$, we compute the timestep embedding $\mathbf{s}(t)$ via sinusoidal positional encoding \cite{supp_sinusoidal_positional_encoding__vaswani2017attention} followed by an MLP. For each modality stream $m \in \{\mathrm{occ}, \mathrm{txt}\}$, a learned MLP predicts two sets of modulation parameters: 
\begin{equation} 
    (\boldsymbol\alpha_{m,1}, \boldsymbol\beta_{m,1}, \mathbf{g}_{m,1}, \boldsymbol\alpha_{m,2}, \boldsymbol\beta_{m,2}, \mathbf{g}_{m,2}) = \mathrm{MLP}_m^{(l)}(\mathbf{s}(t)). 
\end{equation} 
We apply AdaLN to prepare the queries, keys, and values for the attention mechanism: 
\begin{equation} 
    \hat{\mathbf{X}}_m = \mathrm{LN}(\mathbf{X}_m^{(l-1)}) \odot (1 + \boldsymbol\alpha_{m,1}) + \boldsymbol\beta_{m,1}, \quad \text{where } \mathbf{X} \in \{\mathbf{H}, \mathbf{C}\}. 
\end{equation} 

\textbf{(II) Spatial Joint Attention (Dual-Stream).} 
We first model the spatial relationships and text-to-geometry grounding. The modulated occupancy sequence is reshaped into frame-wise batches $\hat{\mathbf{H}}_f \in \mathbb{R}^{S \times d_{\mathrm{model}}}$. 
To encode spatial geometry, we apply 2D Rotary Positional Embeddings (RoPE) to the queries and keys of $\hat{\mathbf{H}}_f$ based on their 2D spatial grid coordinates $(h, w)$. 
\textit{Implementation Note:} Our model conceptually operates on 3D spatial tokens. In our efficient implementation, the $Z$-axis (height) is densely folded into the channel dimension. Thus, applying 2D RoPE on the $H \times W$ grid mathematically serves as the spatial component of the 3D+1D RoPE, preserving full 3D geometric awareness without the cubic complexity of explicit 3D attention.

For each frame $f$, we concatenate the shared text tokens $\hat{\mathbf{C}}$ and the frame-specific occupancy tokens $\hat{\mathbf{H}}_f$ to form a joint sequence. We perform self-attention on this unified sequence to capture cross-modal interactions: 
\begin{equation} 
    [\Delta \mathbf{C}_f; \Delta \mathbf{H}_f] = \mathrm{Attention}([\hat{\mathbf{C}}; \hat{\mathbf{H}}_f]), 
\end{equation} 
where the output is split back into the text update $\Delta \mathbf{C}_f \in \mathbb{R}^{L \times d_{\mathrm{model}}}$ and the spatial occupancy update $\Delta \mathbf{H}_f \in \mathbb{R}^{S \times d_{\mathrm{model}}}$. 

The streams are then updated via residual connections gated by $\mathbf{g}_{m,1}$: 
\begin{align} 
    \mathbf{H}' &= \mathbf{H}^{(l-1)} + \mathbf{g}_{\mathrm{occ},1} \odot \mathrm{Concat}(\Delta \mathbf{H}_{1}, \dots, \Delta \mathbf{H}_{F}), \\ 
    \mathbf{C}' &= \mathbf{C}^{(l-1)} + \mathbf{g}_{\mathrm{txt},1} \odot \frac{1}{F}\sum_{f=1}^{F} \Delta \mathbf{C}_f. 
\end{align} 

\textbf{(III) Temporal Self-Attention (Image Only).} 
Following the spatial interaction, we model the temporal dynamics exclusively on the occupancy stream. We reshape the intermediate tokens $\mathbf{H}'$ to isolate temporal tubes $\mathbf{H}'_s \in \mathbb{R}^{F \times d_{\mathrm{model}}}$ for each spatial token $s \in \{1, \dots, S\}$. 
We inject temporal order via 1D Temporal RoPE based on the frame index $f$. The sequence is updated via a temporal self-attention layer, scaled by a learnable scalar gate $\gamma_{\mathrm{tmp}}$: 
\begin{equation} 
    \mathbf{H}'' = \mathbf{H}' + \gamma_{\mathrm{tmp}} \cdot \mathrm{MHSA}_{\mathrm{tmp}}(\mathbf{H}'_s). 
\end{equation} 

\textbf{(IV) Post-Modulation and MLP.} 
Finally, we process the tokens through the feed-forward networks. We apply the second set of modulation parameters to the updated tokens $\mathbf{H}''$ and $\mathbf{C}'$: 
\begin{align} 
    \tilde{\mathbf{H}} &= \mathrm{LN}(\mathbf{H}'') \odot (1 + \boldsymbol\alpha_{\mathrm{occ},2}) + \boldsymbol\beta_{\mathrm{occ},2}, \\ 
    \tilde{\mathbf{C}} &= \mathrm{LN}(\mathbf{C}') \odot (1 + \boldsymbol\alpha_{\mathrm{txt},2}) + \boldsymbol\beta_{\mathrm{txt},2}. 
\end{align} 
The outputs of the transformer block are obtained by adding the MLP residuals, gated by $\mathbf{g}_{m,2}$: 
\begin{align} 
    \mathbf{H}^{(l)} &= \mathbf{H}'' + \mathbf{g}_{\mathrm{occ},2} \odot \mathrm{MLP}_{\mathrm{occ}}(\tilde{\mathbf{H}}), \\ 
    \mathbf{C}^{(l)} &= \mathbf{C}' + \mathbf{g}_{\mathrm{txt},2} \odot \mathrm{MLP}_{\mathrm{txt}}(\tilde{\mathbf{C}}). 
\end{align} 

\paragraph{3. Interaction-Consistent Forecasting via History Anchoring.}
Let $\mathbf{x}_0 \in \mathbb{R}^{F \times S \times d_{\mathrm{model}}}$ denote the ground-truth clean latent sequence, and $\mathbf{x}_t = (1-t)\mathbf{x}_0 + t\mathbf{x}_1$ denote the noisy sequence at flow step $t \in [0,1]$. 
Given a history horizon of $h$ frames, we define a temporal token replacement operator $\mathcal{R}_{\mathrm{rep}}(\cdot, \cdot)$. For a target sequence $\mathbf{x}_t$ and a clean reference $\mathbf{x}_0$, the operator yields a hybrid sequence $\tilde{\mathbf{x}}_t = \mathcal{R}_{\mathrm{rep}}(\mathbf{x}_t, \mathbf{x}_0)$, where for each frame index $f \in \{1, \dots, F\}$:
\begin{equation}
    \tilde{\mathbf{x}}_{t, f} = 
    \begin{cases} 
    \mathbf{x}_{0, f}, & \text{if } f \le h \text{ (Clean History Anchor at } t=0 \text{)}, \\
    \mathbf{x}_{t, f}, & \text{if } f > h \text{ (Noisy Future to Predict)}.
    \end{cases}
\end{equation}
This operator explicitly anchors the conditioning segment at the \emph{clean timestep} ($t=0$), ensuring the history portion remains uncorrupted and provides a stable spatial-temporal reference.

\textbf{Training Dynamics: Independent Anchoring and CFG.} 
During training, we employ two independent stochastic dropping mechanisms to enable flexible conditioning:
\begin{itemize}
    \item \textit{History Anchoring:} With a probability $p_{\mathrm{anchor}} = 0.5$, we apply the replacement operator to construct the partially clean input $\tilde{\mathbf{x}}_t = \mathcal{R}_{\mathrm{rep}}(\mathbf{x}_t, \mathbf{x}_0)$. Otherwise, the model receives the fully noisy sequence $\mathbf{x}_t$. The loss is only computed on the future frames $(h+1:F)$.
    \item \textit{Classifier-Free Guidance (CFG):} Independently of the anchoring drop, we apply CFG with a probability $p_{\mathrm{cfg}} = 0.15$. When triggered, the refined text tokens $\mathbf{C}$ are replaced by a null embedding sequence $\emptyset$.
\end{itemize}
Because $p_{\mathrm{anchor}}$ and $p_{\mathrm{cfg}}$ are sampled independently, the network $v_\theta$ learns a joint distribution that supports four modes: unconditional generation, text-only generation, history-only forecasting, and joint text-and-history forecasting.

\textbf{Inference via Clean-Timestep Anchoring.} 
During ODE sampling, we integrate backwards from $t=1$ to $t=0$. Let $\hat{\mathbf{x}}_{t-\Delta t}$ be the sequence predicted by the solver at the current step. Given the user-provided clean history context $\mathbf{H}_{1:h}$ (which acts as the ground-truth $\mathbf{x}_0^{(1:h)}$), we forcefully anchor the trajectory at each solver step using the same replacement operator:
\begin{equation}
    \mathbf{x}_{t-\Delta t} \leftarrow \mathcal{R}_{\mathrm{rep}}(\hat{\mathbf{x}}_{t-\Delta t}, \mathbf{H}_{1:h}).
\end{equation}
Because the model was explicitly trained with $\tilde{\mathbf{x}}_t$ via the independent anchoring probability, this step-wise replacement introduces zero distribution shift. It guarantees that the generated future strictly adheres to the provided history context without accumulating integration errors.

\paragraph{Hyperparameters.} 
We summarize the detailed hyperparameters of \ModelName{} in \cref{tab:hyperparams}.

\begin{table}[t]
    \centering
    \small
    \caption{\textbf{Detailed Hyperparameters of \ModelName{}.}}
    \label{tab:hyperparams}
    \begin{tabular}{l|c}
    \toprule
    \textbf{Hyperparameter} & \textbf{Value} \\
    \midrule
    \multicolumn{2}{l}{\textit{Model Architecture}} \\
    Hidden Dimension ($d_{\mathrm{model}}$) & $896$ \\
    Transformer Layers (Depth) & $14$ \\
    Attention Heads & $14$ \\
    Head Dimension & $64$ \\
    Token Refiner Layers ($N_{\mathrm{ref}}$) & $2$ \\
    \midrule
    \multicolumn{2}{l}{\textit{Input Specifications \& Tokenization}} \\
    Original Occupancy Size & $F \times 200 \times 200 \times 16$ \\
    Interpolated Occupancy Size & $F \times 224 \times 224 \times 16$ \\
    Occ-VAE Expansion Ratio ($e$) & $8$  \\
    Occ-VAE Latent Channels ($C$) & $16$ \\
    Latent Spatial Size ($H_{\ell} \times W_{\ell}$) & $28 \times 28$ \\
    Patch Size ($p$) & $2$ \\
    Spatial Tokens per Frame ($S$) & $196$ \\
    \midrule
    \multicolumn{2}{l}{\textit{Training Dynamics}} \\
    Batch Size & $32$ \\
    Learning Rate & $1 \times 10^{-4}$ \\
    Optimizer & AdamW ($\beta_1=0.9, \beta_2=0.95$) \\
    Training Iterations & $100\text{k}$ \\
    Classifier-Free Guidance (CFG) Probability & $0.15$ \\
    Anchoring Probability ($p_{\mathrm{anchor}}$) & $0.5$ \\
    \midrule
    \multicolumn{2}{l}{\textit{Inference Settings}} \\
    ODE Solver & Euler \\
    Sampling Steps (NFE) & $20$ \\
    Classifier-Free Guidance (CFG) Scale & $7.5$ \\
    \bottomrule
    \end{tabular}
\end{table}

\subsection{\DatasetName{} Standardization and Annotation Details}
\label{sec:supp_dataset_standardization_prompts}

Following UniOcc \cite{supp_uniocc__wang2025uniocc}, we adopt a unified occupancy grid of $200\times200\times16$ voxels per frame with a voxel size of $0.4\,\mathrm{m}$, covering a spatial range of $[-40\,\mathrm{m}, 40\,\mathrm{m}]$ for the $X$ and $Y$ axes, and $[-3.2\,\mathrm{m}, 3.2\,\mathrm{m}]$ for the $Z$ axis. Occupancy sequences are sampled at $2\,\mathrm{Hz}$. For all synthetic sources, we map the raw semantic labels to the same 11 categories (including $10$ semantic classes and one empty class) defined by the UniOcc protocol. The detailed mapping from source datasets, as well as several other common autonomous driving benchmarks, to our unified categories is summarized in \cref{tab:label_mapping}.

\begin{table}[t]
    \centering
    \small
    \caption{\textbf{Semantic Category Mapping for \DatasetName{}.}}
    \label{tab:label_mapping}
    \resizebox{\linewidth}{!}{%
    \renewcommand{\arraystretch}{1.3}
    \begin{tabular}{c|>{\raggedright\arraybackslash}p{3cm}|>{\raggedright\arraybackslash}p{3cm}|>{\raggedright\arraybackslash}p{3cm}|>{\raggedright\arraybackslash}p{3cm}}
    \toprule
    \textbf{ID} & \textbf{UniOcc Category (Ours)} \cite{supp_uniocc__wang2025uniocc} & \textbf{nuScenes Category} \cite{supp_occ3d__tian2023occ3d} & \textbf{Waymo Category} \cite{supp_occ3d__tian2023occ3d} & \textbf{Carla Category} \cite{supp_carlasc__wilson2022motionsc} \\
    \midrule
    0 & General Object & General object, Barrier & General object, Pole, Sign & Fences, Other, Poles, Walls, Traffic signs \\ \addlinespace
    1 & Vehicle & Bus, Car, Construction vehicle, Trailer, Truck & Vehicle & Vehicles \\ \addlinespace
    2 & Bicycle & Bicycle & Bicycle & --- \\ \addlinespace
    3 & Motorcycle & Motorcycle & Motorcycle & --- \\ \addlinespace
    4 & Pedestrian & Pedestrian & Pedestrian, Cyclist & Pedestrians \\ \addlinespace
    5 & Traffic Cone & Traffic cone & Traffic light, Construction cone & --- \\ \addlinespace
    6 & Vegetation & Vegetation & Vegetation, Tree trunk & Vegetation \\ \addlinespace
    7 & Road & Drivable surface & Road & Roadlines, Roads \\ \addlinespace
    8 & Walkable/Terrain & Sidewalk, Terrain, Other flat & Walkable & Sidewalks, Ground \\ \addlinespace
    9 & Building & Manmade & Building & Buildings \\ \addlinespace
    10 & Free & Free & Free & Free, Sky \\
    \bottomrule
    \end{tabular}
    }
\end{table}

We detail the prompt engineering process used to automatically generate the language annotations for \DatasetName{}. We employ \texttt{qwen3-vl-235b-a22b-instruct} as our frozen annotator across all data generation pipelines.

Below, we present the complete prompt templates (VLM inputs) and representative output examples for the three sub-datasets:
\begin{itemize}
    \item \textbf{\StaticDatasetName{}:} The prompt is designed to translate 2D semantic BEV projections into ego-centric, 3D spatial layout descriptions, explicitly forbidding the mention of rendering colors.
    \item \textbf{\EnvDatasetName{}:} The prompt requires the VLM to jointly reason over a front-view video sequence and a static BEV road map to deduce ego-vehicle maneuvers and describe the local topology.
    \item \textbf{\AgentDatasetName{}:} The prompt focuses on analyzing multi-agent interactions from video sequences and assigns a criticality score ($0-10$) to enable the risk stratification discussed in the main paper.
\end{itemize}

\vspace{0.5em}


\providecommand{\PromptImagePlaceholder}[2][0.9\linewidth]{%
  \begingroup
  \setlength{\fboxsep}{0pt}%
  \setlength{\fboxrule}{0.6pt}%
  \noindent\fbox{\colorbox{white}{\parbox[c][0.22\textheight][c]{#1}{\centering\scriptsize #2}}}%
  \endgroup
}

\lstdefinestyle{PromptTextStyle}{%
  language={},
  basicstyle=\scriptsize\ttfamily,
  columns=fullflexible,
  keepspaces=true,
  showstringspaces=false,
  breaklines=true,
  breakatwhitespace=true,
  upquote=true
}

%


\tcbset{PromptJsonBox/.style={
    enhanced,
    colframe=gray!40,
    colback=gray!2,
    colbacktitle=gray!15,
    coltitle=black,
    fonttitle=\bfseries\footnotesize,
    boxrule=0.6pt,
    arc=3pt,
    left=4pt, right=4pt, top=3pt, bottom=3pt,
    boxsep=4pt,
    listing only,
    listing options={style=PromptTextStyle},
    breakable,
}}

\tcbset{PromptFullBox/.style={
    boxrule=1pt,
    boxsep=2pt,
    colback=gray!5,
    fontupper=\scriptsize,
    coltitle=black,
    colframe=black,
    colbacktitle=blue!10,
    breakable
}}


\begin{tcolorbox}[PromptFullBox,title=\StaticDatasetName{} (single-frame BEV layouts): Example Input]

\begin{minipage}[t]{0.62\linewidth}
\vspace{0pt}
\begin{lstlisting}[style=PromptTextStyle]
You are an AI assistant tasked with describing driving scenarios. Your goal is to interpret a 2D top-down semantic projection of a 3D scene and generate a natural, flowing, and unstructured English paragraph that describes the underlying 3D scene's spatial layout on the ground plane.

This caption is for training a model to generate the 3D occupancy grid from text, so accuracy in planar relationships is paramount.

**--- SCENE PROJECTION LEGEND ---**

This is the color-coding key for the top-down projection of the 3D scene. Use it to identify the objects and structures.

*   **Road:** Magenta `(255, 0, 255)`
*   **Car:** Blue `(0, 150, 245)`
*   **Pedestrian:** Red `(255, 0, 0)`
*   **Building:** Lavender `(230, 230, 250)`
*   **Vegetation:** Green `(0, 175, 0)`
*   **Terrain:** Dark Gray `(100, 100, 100)` (e.g., sidewalks, curbs, dirt)
*   **Motorcycle:** Dark Orange `(200, 180, 0)`
*   **Bicycle:** Pink `(255, 192, 203)`
*   **Traffic Cone:** Light Yellow `(255, 240, 150)`
*   **Undefined (White):** Represents areas with no sensor data. In your description, refer to these as **"unseen" or "occluded" regions**.
*   **Free (Transparent/Black):** Represents unoccupied drivable space.
\end{lstlisting}
\end{minipage}\hfill
\begin{minipage}[t]{0.35\linewidth}
\vspace{0pt}
\centering
\includegraphics[width=\linewidth]{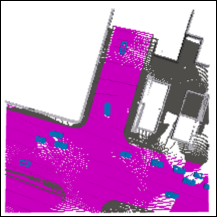}
\end{minipage}

\vspace{0.5ex}

\begin{lstlisting}[style=PromptTextStyle]
**--- CAPTIONING GUIDELINES ---**

Follow these rules precisely to generate the ideal caption:

1.  **Describe Concepts, Not Visuals:** Your task is to describe the real-world scene, not the rendered image.
    *   You **MUST NOT** mention colors or visual appearance details from the rendering.
    *   **Correct:** `The scene contains a road and a car.`
    *   **Incorrect:** `The scene has a magenta road and a blue car.`

2.  **Use Accurate Ego-Centric Directions:** All descriptions must be from the perspective of the ego-vehicle, located at the center of the image (indicated by a dark blue square) and facing forward.
    *   Image **Top** = Scene **Front / Ahead**
    *   Image **Bottom** = Scene **Rear / Behind**
    *   **Correct:** `A car is located to the rear-right.`
    *   **Incorrect:** `A car is in the bottom-right corner of the image.`

3.  **Describe Static State, Not Dynamics:** You are describing a single snapshot in time. **Do not infer motion, speed, or traffic flow.**
    *   Describe object placement and density, not their movement. Use terms like 'density', 'distribution', 'placement', and 'position'.
    *   **AVOID** words like 'moving', 'stopped', 'stationary', 'traffic flow', 'overtaking'.
    *   **Correct:** `The density of vehicles ahead is high.`
    *   **Incorrect:** `The traffic flow ahead is heavy, and the cars are stationary.`

4.  **Strictly Omit 3D Vertical Information:** Your description must only contain information about the layout on the ground (X-Y plane). **Do not** mention or infer object height, volume, or elevation.
    *   **Correct:** `There is a building on the right.`
    *   **Incorrect:** `There is a tall building on the right.`

5.  **Describe "Undefined" as "Unseen":** The white regions represent areas outside the sensor's field of view or areas occluded by other objects. Describe them naturally as **"unseen areas"**, **"occluded regions"**, or **"areas beyond sensor range"**.

---

**Output Format:**
Return the result strictly in JSON format without Markdown code block encapsulation.

```json
{
    "scene_caption": "A description of the 3D driving scene based on the BEV projection."
}
```

Now, generate a caption for the BEV scene projection in the image, strictly adhering to all guidelines and the specified multi-paragraph format.
\end{lstlisting}

\end{tcolorbox}

\begin{tcolorbox}[PromptJsonBox,title=\StaticDatasetName{}: Example Output]
\begin{lstlisting}[style=PromptTextStyle]
{
  "scene_caption": "A scene of a multi-lane road with a complex intersection or merge point ahead, situated in a dense urban environment. "
}
\end{lstlisting}
\end{tcolorbox}

\vspace{8pt}
\begin{tcolorbox}[PromptFullBox,title=\EnvDatasetName{} (turn-point maneuver clips): Example Input]

\begin{center}
\includegraphics[width=\linewidth]{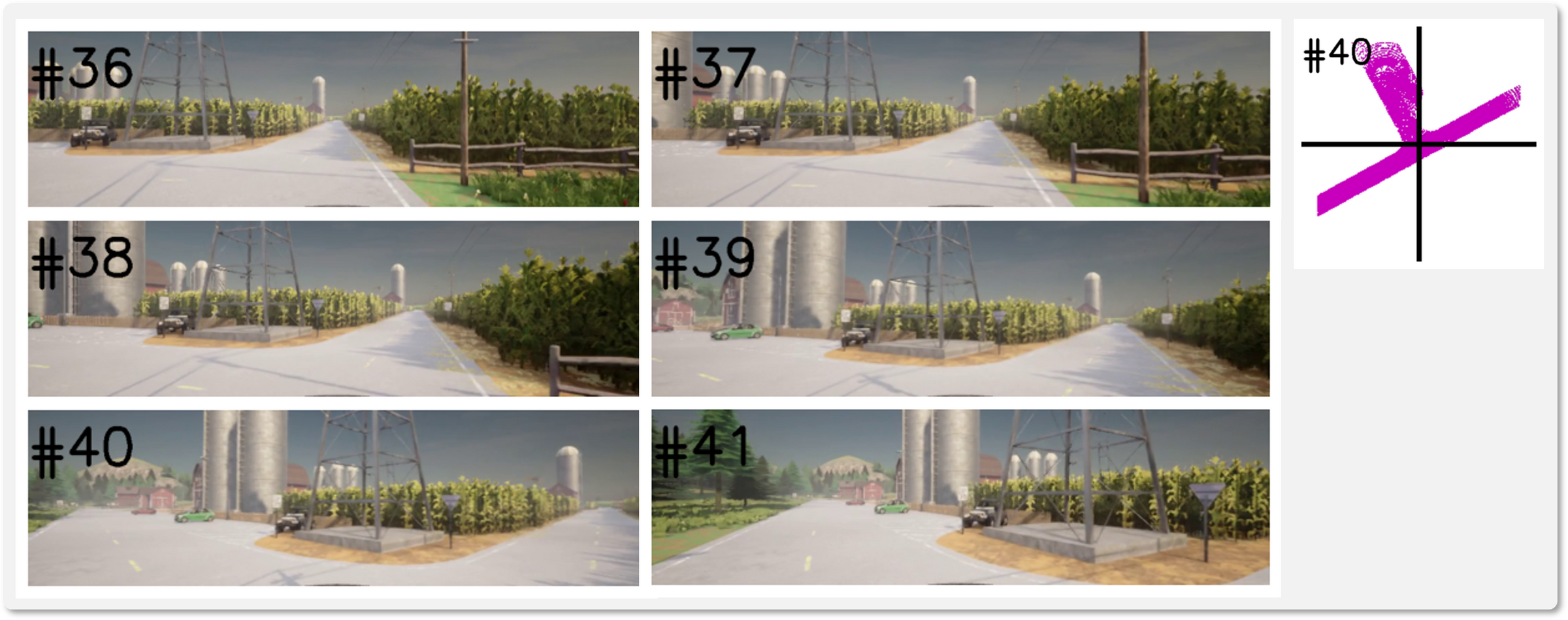}
\end{center}

\vspace{1ex}

\begin{lstlisting}[style=PromptTextStyle]
**Role:**
You are an expert Image Analyst specialized in autonomous driving scenarios. You will receive a sequence of real world (or simulation) driving video frames captured from the ego vehicle's front windshield, followed by a final reference image.

**Task:**
Analyze the ego vehicle's temporal driving behavior and the surrounding spatial environment based on the visual sequence and the provided road structure map.

**Input Data Description:**

1.  **Image Sequence:**
    *   The images are arranged in strictly ascending chronological order.
    *   **Frame IDs:** Displayed as black numbers in the top-left corner. Use these to track temporal progression.

2.  **Road Structure Map (Bird's-Eye View):**
    *   **The Final Image:** The very last image in the input list is a BEV (Bird's-Eye View) map representing the static road structure around the ego vehicle.
    *   **Temporal Alignment:** The Frame ID in the top-left of this map corresponds to the frame in the video sequence where this road structure applies.
    *   **Ego Vehicle Positioning:** Two perpendicular black lines indicate the ego vehicle's pose:
        *   **Vertical Line:** Represents the vehicle's heading (pointing Up).
        *   **Horizontal Line:** Represents the vehicle's lateral axis (Left/Right).
        *   **Intersection:** The ego vehicle is located exactly at the intersection of these two lines.
    *   **Semantic Colors:** The map is simplified. Only the traversable road surface is highlighted in **Magenta (255, 0, 255)**.

---

**Guidelines**

1.  **Behavior Classification (Taxonomy):**
    Determine the ego vehicle's action based on the sequence:
    *   **STRAIGHT:** Vehicle proceeds forward without significant deviation.
    *   **LEFT:** Vehicle executes a left turn.
    *   **RIGHT:** Vehicle executes a right turn.

2.  **Environmental Description:**
    Generate a description of the driving environment. Focus on spatial layout, surrounding structures (buildings), vegetation, and road morphology (geometry/shape).

3. **Critical Constraints (For Occupancy Labeling):**
    The output description will be used as a text label for a Occupancy  training pipeline. Therefore, you must adhere to the following:
    *   **Ignore Simulation Artifacts:** Even if the images appear to be from a simulator, describe the scene as if it were the real world. Do not mention "simulation," "game engine," or "synthetic data."
    *   **Focus on Geometry/Structure:** Do not include details imperceptible to a geometric occupancy sensor (e.g., specific weather conditions, color, texture details, or lighting). Focus on *what* objects are present and *where* they are.

---

**Output Format:**
Return the result strictly in JSON format without Markdown code block encapsulation.

```json
{
    "action": "STRAIGHT | LEFT | RIGHT",
    "description": "A concise summary of the ego vehicle's behavior, including the shape and width of the road."
}
```
\end{lstlisting}

\end{tcolorbox}

\begin{tcolorbox}[PromptJsonBox,title=\EnvDatasetName{}: Example Output]
\begin{lstlisting}[style=PromptTextStyle]
{
    "action": "LEFT",
    "description": "The ego vehicle executes a left turn at a Y-intersection onto a two-lane road. The road is flanked by fields of tall crops and industrial structures including silos."
}
\end{lstlisting}
\end{tcolorbox}

\vspace{8pt}
\begin{tcolorbox}[PromptFullBox,title=\AgentDatasetName{} (multi-agent interactions): Example Input]

\begin{center}
\includegraphics[width=\linewidth]{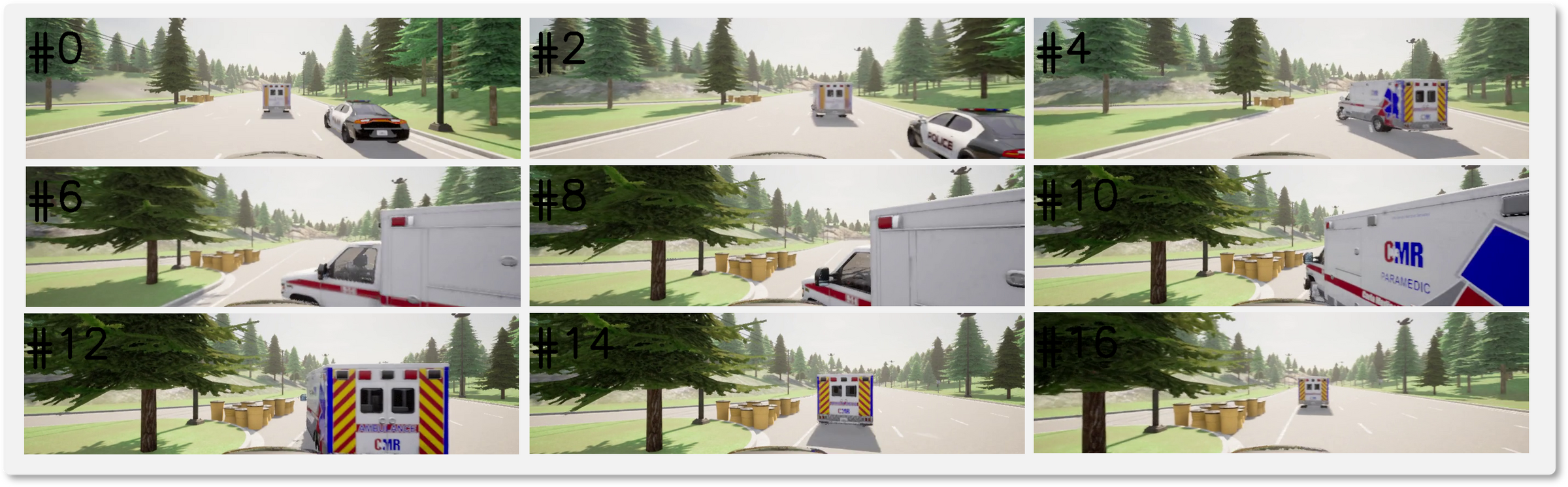}
\end{center}

\vspace{1ex}

\begin{lstlisting}[style=PromptTextStyle]
**Role:**
You are an expert Image Analyst specialized in autonomous driving scenarios. You will receive a sequence of real world (or simulation) driving video frames captured from the ego vehicle's front windshield.

**Task:**
Analyze the driving behavior, environment, and potential risks based on the provided frames.

**Input Data Description:**
1.  **Image Sequence:** The images are arranged in ascending chronological order.
2.  **Meta Data:** Frame IDs are displayed as Black Numbers in the top-left corner (use these to understand the temporal progression).

**Guidelines:**

1.  **Environment Analysis:** Analyze the situation in the sequence. Describe the environment, including surrounding vehicles, pedestrians, buildings, vegetation, road types.

2.  **Safety Scoring:** Evaluate the necessity for the ego vehicle (driver) to exercise caution or take emergency action. Assign a **score (0-10)** based on the following scale:
    *   **0-2:** Safe driving conditions. Good visibility, no immediate hazards (e.g., clear road).
    *   **3-5:** Caution required. Vulnerable road users (pedestrians, cyclists) or potential obstacles present in the field of view.
    *   **6-8:** Immediate action/Emergency required. High-risk situations (e.g., vehicle cutting in, sudden braking ahead, jaywalking).
    *   **9-10:** Collision inevitable or has occurred.

3.  **Critical Constraints (For Occupancy Labeling):**
    The output description will be used as a text label for a Occupancy  training pipeline. Therefore, you must adhere to the following:
    *   **Ignore Simulation Artifacts:** Even if the images appear to be from a simulator, describe the scene as if it were the real world. Do not mention "simulation," "game engine," or "synthetic data."
    *   **Focus on Geometry/Structure:** Do not include details imperceptible to a geometric occupancy sensor (e.g., specific weather conditions, color, texture details, or lighting). Focus on *what* objects are present and *where* they are.

**Output Format:**
Return the result strictly in JSON format. Do not output markdown code blocks alone, just the raw JSON string.

```json
{
    "description": "Briefly describe the ego vehicle's driving behavior and the cause of risks (if any).",
    "score": <Integer between 0-10>
}
```
\end{lstlisting}

\end{tcolorbox}

\begin{tcolorbox}[PromptJsonBox,title=\AgentDatasetName{}: Example Output]
\begin{lstlisting}[style=PromptTextStyle]
{
  "description_short": "The ego vehicle follows an ambulance which suddenly spins out and blocks the road, forcing the ego vehicle to take emergency evasive action.",
  "score": 8
}
\end{lstlisting}
\end{tcolorbox}

\subsection{Prompt Templates for Automated Quality Assessment}
\label{sec:supp_dataset_quality_prompts}

We provide the complete rubric-based judge prompt used in ``Automated Quality Assessment''. The prompt is applied to a BEV image sequence rendered from the semantic occupancy grid and a text description (the caption or generation prompt). Below, we present the system message defining the evaluation rubric and a representative output example generated by the VLM judge.

\vspace{0.5em}


\providecommand{\PromptImagePlaceholder}[2][0.9\linewidth]{%
  \begingroup
  \setlength{\fboxsep}{0pt}%
  \setlength{\fboxrule}{0.6pt}%
  \noindent\fbox{\colorbox{white}{\parbox[c][0.22\textheight][c]{#1}{\centering\scriptsize #2}}}%
  \endgroup
}

\lstdefinestyle{PromptTextStyle}{%
  language={},
  basicstyle=\scriptsize\ttfamily,
  columns=fullflexible,
  keepspaces=true,
  showstringspaces=false,
  breaklines=true,
  breakatwhitespace=true,
  upquote=true
}

%


\tcbset{PromptJsonBox/.style={
    enhanced,
    colframe=gray!40,
    colback=gray!2,
    colbacktitle=gray!15,
    coltitle=black,
    fonttitle=\bfseries\footnotesize,
    boxrule=0.6pt,
    arc=3pt,
    left=4pt, right=4pt, top=3pt, bottom=3pt,
    boxsep=4pt,
    listing only,
    listing options={style=PromptTextStyle},
    breakable,
}}

\tcbset{PromptFullBox/.style={
    boxrule=1pt,
    boxsep=2pt,
    colback=gray!5,
    fontupper=\scriptsize,
    coltitle=black,
    colframe=black,
    colbacktitle=blue!10,
    breakable
}}


\begin{tcolorbox}[PromptFullBox,title=Automated Quality Assessment: Example Input]

\begin{center}
\includegraphics[width=\linewidth]{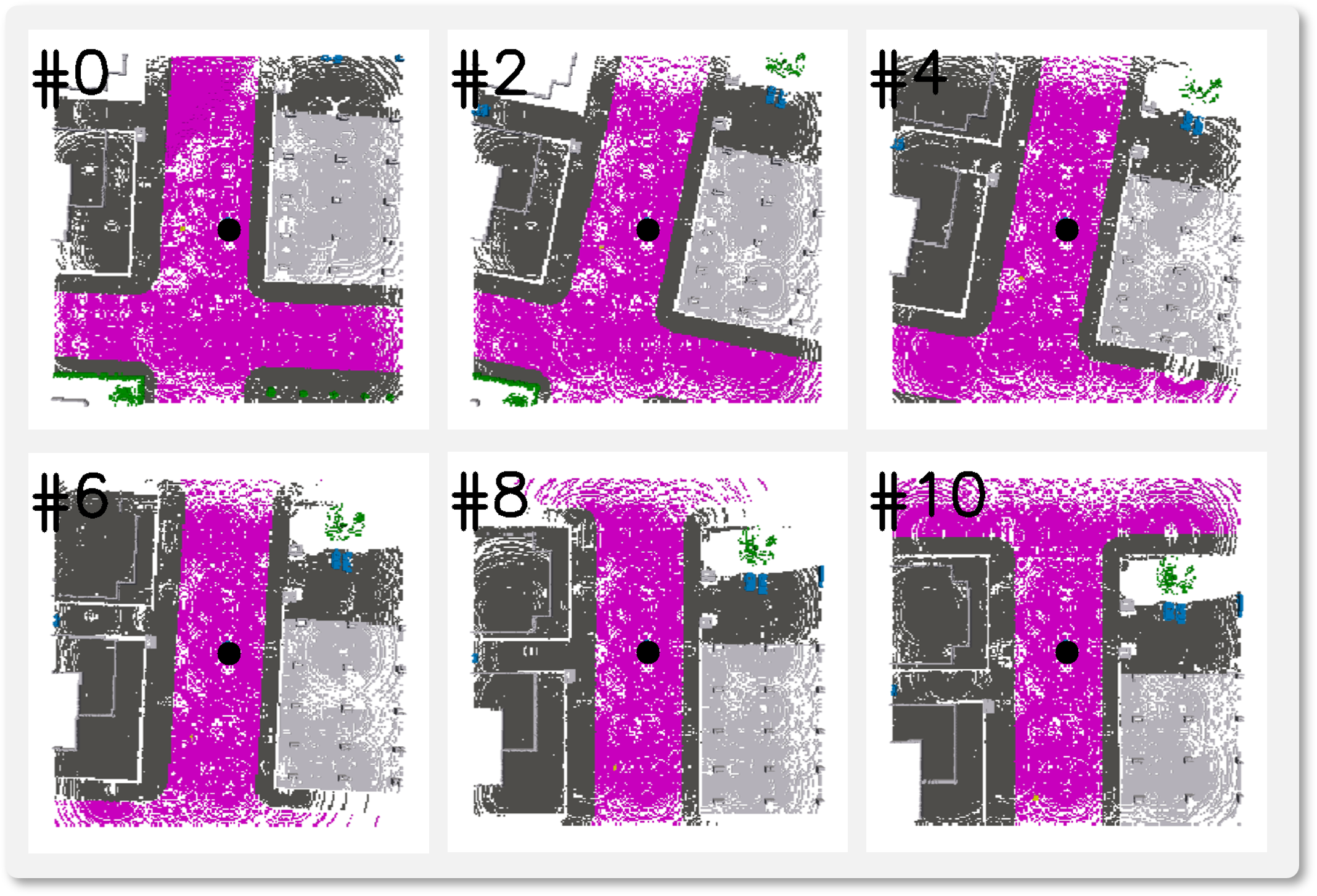}
\end{center}

**Related prompt:** The scene is an urban environment with multi-lane roads and tall, multi-story buildings lining both sides. The ego vehicle is positioned in the center lane and proceeds straight ahead. No other vehicles, pedestrians, or moving objects are visible in the field of view at any point in the sequence. The road markings are clear, and the path is unobstructed. The surrounding infrastructure includes streetlights, a low wall on the right, and some small structures on the sidewalks. The overall situation is static and poses no immediate threat.

\vspace{1ex}

\begin{lstlisting}[style=PromptTextStyle]
You are an expert evaluator for an autonomous driving scene generation system. Your task is to score generated Bird's-Eye View (BEV) driving video frames based on three criteria: Scene Completeness, Structural Correctness, and Instruction Following.

**Input Data Description:**

(1) A text Prompt that was used to generate the scene.

(2) Bird's-Eye View Image Sequence (video frames):
    - Images are in strictly ascending chronological order.
    - Frame IDs are shown as black numbers in the top-left corner for tracking temporal progression.

**Scene projection legend (BEV semantic colors):**
- Road: Magenta (255, 0, 255)
- Car: Blue (0, 150, 245)
- Pedestrian: Red (255, 0, 0)
- Building: Lavender (230, 230, 250)
- Vegetation: Green (0, 175, 0)
- Terrain: Dark Gray (100, 100, 100)
- Undefined (White): unseen/empty areas where the model generated nothing.
- Free (Transparent/Black): unoccupied drivable space.

**Ego vehicle positioning:** a black point at the center indicates the ego pose; the area above it is the front of the vehicle.

**Evaluation criteria & scoring rubrics (1-5 points):**

1) Scene Completeness. Assess how much of the scene is filled versus being empty (white/undefined), ignoring image edges.
    - 1 (Bad): almost entirely white/empty; failed to generate meaningful content.
    - 2 (Poor): large unnatural white patches dominate the scene.
    - 3 (Fair): noticeable empty areas that hurt realism.
    - 4 (Good): mostly complete; white areas are minimal and acceptable.
    - 5 (Excellent): fully rendered without unnatural empty patches.

2) Structural Correctness. Assess physical plausibility and topological consistency independent of the prompt.
    - 1 (Bad): nonsensical layout (e.g., buildings on roads, chaotic placement).
    - 2 (Poor): major flaws (illogical breaks, strange relationships).
    - 3 (Fair): plausible but contains minor unrealistic elements.
    - 4 (Good): logical and consistent driving environment.
    - 5 (Excellent): highly realistic and coherent complex layout.

3) Instruction Following. (Most important) Assess how well the BEV sequence matches the given prompt.
    - 1 (Bad): ignores/contradicts the prompt.
    - 2 (Poor): captures only a minor element; misses the main theme.
    - 3 (Fair): matches general theme but misses key details.
    - 4 (Good): generates the described scene and primary objects; may miss minor nuances.
    - 5 (Excellent): perfect match, including specified elements and spatial relations.

**Additional instructions:** ignore any color information mentioned in the prompt (e.g., traffic light state or vehicle paint color), since the BEV semantic rendering uses a fixed color legend.

**Output format:** You must respond with a single JSON object and no surrounding text.

{
  "thought": "Brief rationale for the scores.",
  "scores": {
    "completeness_score": 1-5,
    "structural_correctness_score": 1-5,
    "instruction_following_score": 1-5
  }
}

Now, evaluate the provided BEV image sequence based on the prompt.
\end{lstlisting}
\end{tcolorbox}

\begin{tcolorbox}[PromptJsonBox,title=Automated Quality Assessment: Example Output]
\begin{lstlisting}[style=PromptTextStyle]
{
  "thought": "1. Scene Completeness: The generated scene is well-populated. The road, terrain, and building areas are clearly defined across all frames. While there is some minor 'salt-and-pepper' noise within the semantic regions (common in occupancy grid projections), there are no large, unnatural empty patches that disrupt the scene's realism. Score: 4.
  2. Structural Correctness: The scene exhibits excellent temporal and spatial consistency. As the ego vehicle moves forward (indicated by the frame sequence), the environment scrolls backward smoothly and logically. The road geometry, transitioning from a straight segment to a T-intersection, is physically plausible and represents a realistic urban layout. Score: 5.
  3. Instruction Following: The image follows the core instruction of generating an urban environment where the ego vehicle proceeds straight ahead on an unobstructed path. However, it fails on specific constraints. The prompt explicitly states, \"No other vehicles, pedestrians, or moving objects are visible,\" yet there are distinct blue blocks (cars) visible on the right side (likely parked) and a persistent red speck (pedestrian) in the road area in multiple frames. Additionally, the prompt describes buildings \"lining both sides,\" but the semantic segmentation shows buildings (lavender) primarily on the right, while the left side is dominated by terrain (dark gray). Score: 3.",
  "scores": {
    "completeness_score": 4,
    "structural_correctness_score": 5,
    "instruction_following_score": 3
  }
}
\end{lstlisting}
\end{tcolorbox}

\subsection{Evaluation Protocol and Metric Definitions}
\label{sec:supp_eval_protocol}

\paragraph{Rendering occupancy to BEV RGB.}
Following previous work \cite{supp_semcity__lee2024semcity,supp_dynamiccity__bian2024dynamiccity,supp_X-scene__yang2025x}, we render the 3D/4D occupancy grids into BEV-view RGB images and clips. These rendered RGB representations are subsequently utilized to evaluate all single-frame and video-level visual quality metrics.

\paragraph{Data preparation for visual metrics.}
To evaluate the visual quality and distributional realism of the generated samples, we construct a standardized evaluation set. Specifically, we uniformly sample 10K ground-truth 16-frame occupancy clips from the dataset. For each evaluated method, we generate 10K corresponding 16-frame clips using the classifier-free guidance (CFG) unconditional branch. For single-frame metrics, we uniformly sample 5 frames from each clip, whereas video-level metrics are computed using the full 16-frame sequences.

\paragraph{Single-frame visual quality.}
Using the sampled 5 frames for each clip, we compute Inception Score (IS) \cite{supp_InceptionScore__barratt2018note}, Fr\'echet Inception Distance (FID) \cite{supp_fid__heusel2017gans}, and Kernel Inception Distance (KID) \cite{supp_kid_binkowski2018demystifying} to assess per-frame fidelity. We also report image-level Precision and Recall computed via pre-trained Inception-v3 \cite{supp_Inception-v3__szegedy2016rethinking} features to explicitly disentangle sample fidelity and distributional diversity.

\paragraph{Video realism and diversity.}
To evaluate temporal coherence and dynamic realism, we compute the Fr\'echet Video Distance (FVD) \cite{supp_fvd_unterthiner2018towards} between the generated and ground-truth 16-frame video distributions. We also report video-level Precision and Recall computed using I3D \cite{supp_i3d__carreira2017quo} features to measure spatio-temporal fidelity and diversity.

\paragraph{VLM-based instruction alignment score.}
We adopt the rubric-based VLM evaluator in ``Automated Quality Assessment'' to quantify instruction alignment of generated sequences. The evaluator assigns integer scores in $[1,5]$ (higher is better) along three axes: scene completeness, structural correctness, and instruction following, and we report their mean. To ensure consistent and rigorous comparison, we fix \texttt{gemini-3.0-pro} as our sole judge VLM for all experiments. We select this model specifically for its high sensitivity to the \emph{Instr.~follow} axis, which is the primary focus of our text-guided generation task, ensuring that the scores faithfully reflect the alignment between prompts and generated 4D occupancies.

\section{Additional Quantitative Results}
\label{sec:supp_quant}

\subsection{Inference Resource Consumption}
\label{sec:supp_efficiency}
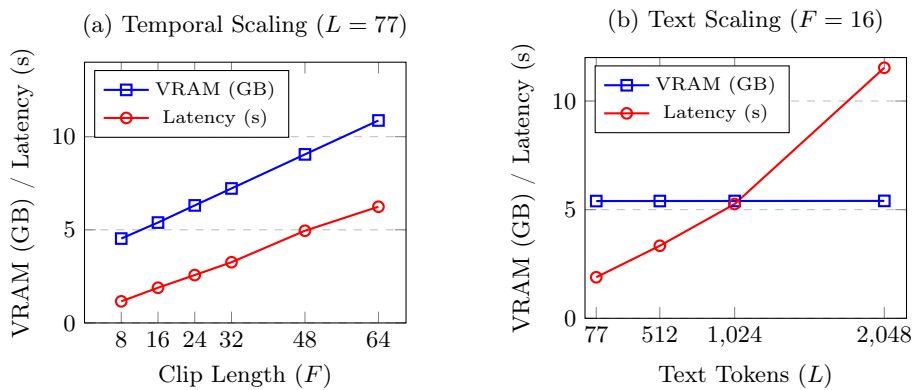
\begin{figure}[t]
  \centering
  \begin{tikzpicture}
    \begin{axis}[
        width=0.48\textwidth,
        xlabel={Clip Length ($F$)},
        ylabel={VRAM (GB) / Latency (s)},
        xmin=0, xmax=70,
        ymin=0, ymax=14,
        xtick={8,16,24,32,48,64},
        legend pos=north west,
        legend style={font=\scriptsize},
        ymajorgrids=true,
        grid style=dashed,
        title={(a) Temporal Scaling ($L=77$)}
    ]
    \addplot[color=blue, mark=square, thick] coordinates {(8,4.53)(16,5.39)(24,6.31)(32,7.22)(48,9.05)(64,10.87)};
    \addlegendentry{VRAM (GB)}
    \addplot[color=red, mark=o, thick] coordinates {(8,1.16)(16,1.89)(24,2.57)(32,3.26)(48,4.95)(64,6.24)};
    \addlegendentry{Latency (s)}
    \end{axis}
  \end{tikzpicture}
  \hfill
  \begin{tikzpicture}
    \begin{axis}[
        width=0.48\textwidth,
        xlabel={Text Tokens ($L$)},
        ylabel={VRAM (GB) / Latency (s)},
        xmin=0, xmax=2200,
        ymin=0, ymax=12,
        xtick={77,512,1024,2048},
        legend pos=north west,
        legend style={font=\scriptsize},
        ymajorgrids=true,
        grid style=dashed,
        title={(b) Text Scaling ($F=16$)}
    ]
    \addplot[color=blue, mark=square, thick] coordinates {(77,5.397)(512,5.398)(1024,5.400)(2048,5.404)};
    \addlegendentry{VRAM (GB)}
    \addplot[color=red, mark=o, thick] coordinates {(77,1.89)(512,3.34)(1024,5.26)(2048,11.53)};
    \addlegendentry{Latency (s)}
    \end{axis}
  \end{tikzpicture}
  \caption{\textbf{Inference resource scaling.} (a) VRAM and latency scale linearly with frames $F$. (b) VRAM is invariant to text length $L$ as the memory peak is dominated by the occupancy stream's temporal attention. Measurements on NVIDIA A100 (80GB) using 20 Euler steps.}
  \label{fig:supp_efficiency}
\end{figure}

We evaluate the inference efficiency of \ModelName{} across varying output temporal lengths ($F$) and input text token counts ($L$). As shown in Fig.~\ref{fig:supp_efficiency}, our Spatio-Temporal Separated Attention (STSA) maintains a strictly linear memory growth relative to clip length. Notably, the peak VRAM remains remarkably invariant to the text token length $L$ during 4D generation ($F \ge 16$).

This memory robustness stems from our decoupled attention design: the memory-intensive temporal attention blocks exclusively process the occupancy stream, while the cross-modal spatial interactions (where $L$ is involved) consume significantly less memory than the temporal bottleneck. Consequently, increasing $L$ from 77 to 2048 yields a negligible VRAM increase ($< 0.1\%$), enabling \ModelName{} to handle highly complex behavioral instructions without additional memory overhead.

\subsection{Architectural Sensitivity}
\label{sec:supp_sensitivity}
We investigate the impact of backbone depth, refiner complexity, and text encoder selection on model performance.

\paragraph{Backbone and Refiner Depth.} 
As shown in Table~\ref{tab:hyper_sensitivity}, increasing MMDiT layers from 10 to 14 yields consistent gains in FVD, confirming that larger model capacity is beneficial for capturing complex 4D dynamics. For the Token Refiner, we observe that a depth of 2 layers is sufficient; further increasing it to 4 layers yields diminishing returns, suggesting that the bottleneck lies in the generative backbone rather than the alignment bridge.

\paragraph{VLM Type.}
We compare different VLM encoders in Table~\ref{tab:supp_vlm_type}. Our default Qwen3-VL-2B refiner achieves the best FVD and Mean VLM Score, even outperforming the larger Qwen2.5-VL-7B. This suggests that the improved visual-linguistic grounding capabilities of the latest VLM generation are more critical for 4D occupancy alignment than raw parameter count.

\begin{table}[t]
\centering
\small
\caption{\textbf{Architectural Sensitivity.} We report FVD and Mean VLM score for various model depths and text encoders.}
\label{tab:hyper_sensitivity}
\begin{tabular}{@{}lcc|cc@{}}
\toprule
Variant & MMDiT Layers & Refiner Depth & FVD $\downarrow$ & VLM Mean $\uparrow$ \\ \midrule
Small        & 10           & 1             & 512            & 2.12            \\
Base         & 12           & 2             & 339            & 3.18            \\
Large (Ref1) & 14           & 1             & 298            & 3.42            \\
Large (Ours) & 14           & 2             & 275   & \textbf{3.73}   \\
Large (Ref4) & 14           & 4             & \textbf{274}    & 3.72            \\ \bottomrule
\end{tabular}
\end{table}

\begin{table}[t]
\centering
\small
\caption{\textbf{Comparison of different VLM encoders.} Performance is evaluated on the full \DatasetName{} validation set.}
\label{tab:supp_vlm_type}
\begin{tabular}{@{}lccc@{}}
\toprule
VLM Encoder Type & FVD $\downarrow$ & Semantic Align. $\uparrow$ & Mean VLM Score $\uparrow$ \\ \midrule
Qwen2.5-VL-7B & 282 & \textbf{3.33} & 3.69 \\
Qwen2.5-VL-3B & 315 & 3.02 & 3.35 \\
Qwen3-VL-2B (Ours) & \textbf{275} & 3.32 & \textbf{3.73} \\ \bottomrule
\end{tabular}
\end{table}

\subsection{Methodological Ablations}
\label{sec:supp_more_ablations}

\paragraph{History-Prefix Anchoring Ratio.}
The anchoring ratio $\gamma$ controls the guidance strength from the past during the flow-matching process. Table~\ref{tab:supp_anchoring} shows that $\gamma=0.50$ provides the optimal balance. A lower ratio leads to temporal drifting (higher FVD), while an excessively high ratio ($\gamma=0.75$) slightly degrades both visual quality and semantic alignment as the model becomes over-constrained by the history prefix.

\begin{table}[t]
\centering
\small
\caption{\textbf{Ablation on History-Prefix Anchoring Ratio $\gamma$.}}
\label{tab:supp_anchoring}
\begin{tabular}{@{}lcccc@{}}
\toprule
Ratio $\gamma$ & 0.0 (None) & 0.25 & 0.50 (Ours) & 0.75 \\ \midrule
FVD $\downarrow$ & 354 & 311 & \textbf{275} & 288 \\
VLM Mean $\uparrow$ & 3.43 & 3.55 & \textbf{3.73} & 3.68 \\ \bottomrule
\end{tabular}
\end{table}

\paragraph{Sensitivity to CFG Scale across Sub-datasets.}
We analyze the instruction alignment (VLM Mean) across three sub-corpora under different Classifier-Free Guidance (CFG) scales. As shown in Table~\ref{tab:supp_cfg}, different tasks exhibit varying sensitivities to CFG. Dynamic scenes (\EnvDatasetName{} and \AgentDatasetName{}) benefit from a higher CFG scale ($s=7.5$) to ensure precise spatio-temporal coordination, whereas \StaticDatasetName{} performance peaks earlier at $s=5.0$ and declines as the guidance strength further increases.

\begin{table}[t]
\centering
\small
\caption{\textbf{CFG scale ($s$) sensitivity per sub-dataset.} We report the VLM Mean score.}
\label{tab:supp_cfg}
\begin{tabular}{@{}lcccc@{}}
\toprule
Sub-dataset & $s=1.0$ (None) & $s=5.0$ & $s=7.5$ (Ours) & $s=10.0$ \\ \midrule
\StaticDatasetName{} & 2.52 & \textbf{3.72} & 3.55 & 3.34 \\
\EnvDatasetName{}   & 2.24 & 3.45 & \textbf{4.02} & 3.51 \\
\AgentDatasetName{} & 1.85 & 3.10 & \textbf{3.62} & 3.25 \\ \midrule
Mean $\uparrow$ & 2.20 & 3.42 & \textbf{3.73} & 3.37 \\ \bottomrule
\end{tabular}
\end{table}

\subsection{Robustness of VLM-based Evaluation}
\label{sec:supp_vlm_robustness}

To further validate the objectivity of the instruction alignment scores, we extend the evaluation conducted in the main experiments to include two additional state-of-the-art VLMs: \texttt{gpt-5.2} and \texttt{qwen3-vl-235b-a22b-instruct}. As discussed in the dataset validation section, while different models exhibit systematic scoring offsets, the relative performance trend across methods remains highly consistent.

\begin{table}[t]
\centering
\small
\caption{\textbf{Cross-VLM Evaluation Robustness.} We report the Mean VLM score across different judge models. Despite the variance in absolute values, the performance gain of \ModelName{} remains consistent across all evaluators.}
\label{tab:supp_cross_vlm}
\begin{tabular}{@{}lcccc@{}}
\toprule
Judge VLM & \begin{tabular}[c]{@{}c@{}} DynamicCity \\ + CLIP \end{tabular} & \begin{tabular}[c]{@{}c@{}} DynamicCity \\ + T5 \end{tabular} & \begin{tabular}[c]{@{}c@{}} \ModelName{} \\ (w/o HP) \end{tabular} & \begin{tabular}[c]{@{}c@{}} \ModelName{} \\ (Full) \end{tabular} \\ \midrule
\texttt{gemini-3.0-pro} (Main) & 3.09 & 2.97 & 3.43 & \textbf{3.73} \\
\texttt{gpt-5.2} (Strict) & 2.18 & 2.11 & 2.55 & \textbf{2.82} \\
\texttt{qwen3-vl-235b-a22b-instruct} & 2.58 & 2.44 & 2.92 & \textbf{3.24} \\ \bottomrule
\end{tabular}
\end{table}

As shown in Table~\ref{tab:supp_cross_vlm}, \ModelName{} consistently outperforms all text-adapted baselines across all three judges. Notably, even under the strictest evaluator (\texttt{gpt-5.2}), our full model maintains a significant margin over the strongest baseline (2.82 vs. 2.18), confirming that the improvements in instruction alignment are not model-specific but represent a general advancement in text-to-4D occupancy synthesis.

\subsection{User Study}
\label{sec:supp_user_study}
Following OccSora~\cite{supp_occsora__wang2024occsora} and DynamicCity~\cite{supp_dynamiccity__bian2024dynamiccity}, we conduct a user study to validate our VLM-based automated evaluation. The study includes 30 samples in total, with 10 samples randomly drawn from each method. We recruited 10 human participants to rate the clips. Each participant was asked to rate the generation clips on a 1--5 Likert scale following the same rubric as our VLM scorer (\ie, scene completeness, structural correctness, and semantic alignment).

\begin{table}[t]
\centering
\small
\caption{\textbf{User study results and Correlation.} Human scores are averaged across participants. $\rho$ denotes the Pearson correlation coefficient between human and VLM scores.}
\label{tab:supp_user_study}
\begin{tabular}{@{}lccc|c@{}}
\toprule
Metric & \begin{tabular}[c]{@{}c@{}} DynamicCity + CLIP \end{tabular} & \begin{tabular}[c]{@{}c@{}}\ModelName{}\\ (w/o HP)\end{tabular} & \begin{tabular}[c]{@{}c@{}}\ModelName{}\\ (Full)\end{tabular} & Correlation $\rho$ \\ \midrule
Completeness & 2.85 & 3.45 & \textbf{3.78} & 0.88 \\
Structural & 3.24 & 3.62 & \textbf{3.95} & 0.85 \\
Sem. Align. & 2.55 & 3.10 & \textbf{3.82} & 0.91 \\ \midrule
\textbf{Mean} & 2.88 & 3.39 & \textbf{3.85} & \textbf{0.88} \\ \bottomrule
\end{tabular}
\end{table}

The results in Table~\ref{tab:supp_user_study} confirm that \ModelName{} is strongly preferred by human evaluators. Most importantly, the high Pearson correlation coefficient ($\rho=0.88$) between human ratings and our automated VLM scores justifies the reliability of the latter as a primary evaluation metric for large-scale 4D occupancy generation.

\section{Additional Qualitative Results}
\label{sec:supp_qual}

In this section, we provide a deeper interpretation of the qualitative results, demonstrating how \ModelName{} effectively bridges the semantic-spatiotemporal gap. In all following figures, the ego vehicle is marked with a \textit{black car icon}. Specific visual highlights, such as newly appearing hazards, pedestrians, or intended trajectories, are indicated with \textit{red circles} or \textit{red dashed lines}.

\subsection{Dataset Samples and Criticality Analysis}
\label{sec:supp_dataset_samples}

\begin{figure}[t]
    \centering
    \includegraphics[width=\linewidth]{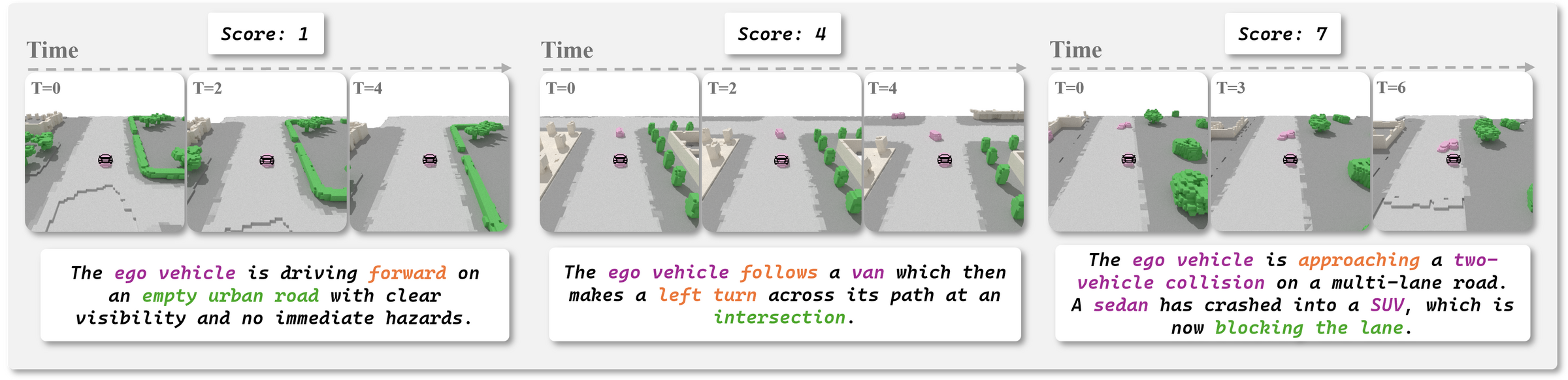}
    \caption{\textbf{Qualitative samples from \AgentDatasetName{} across criticality score regimes.} The samples progress from routine driving (left) to potential conflicts (middle) and imminent hazards (right), demonstrating the breadth of safety-critical scenarios covered by our automated data engine.}
    \label{fig:supp_criticality_samples}
\end{figure}

We visualize representative samples from the \AgentDatasetName{} corpus in \cref{fig:supp_criticality_samples}, categorized by their VLM-predicted criticality scores $s \in [0, 10]$. The samples demonstrate the dataset's ability to cover the full safety-critical spectrum, which is essential for providing diverse supervision during training.

\begin{itemize}
    \item \textbf{Low Criticality (Score 1):} These samples represent routine driving on urban roads. The dataset captures stable 4D occupancy volumes for the ego vehicle and surrounding static elements, providing high-fidelity geometric priors for temporally consistent and physically plausible scenes in low-conflict environments.
    \item \textbf{Medium Criticality (Score 4):} In these scenarios, the corpus contains potential conflicts that require behavioral adaptation. For example, when a lead van executes a left turn across the ego's path, the captured occupancy sequence and corresponding caption accurately reflect the interactive dynamics and the necessary spatial buffers.
    \item \textbf{High Criticality (Score 7):} These cases highlight imminent hazards within the training data. In the crash scenario, the dataset accurately captures the irregular and complex occupancy volumes of the colliding vehicles and the subsequent lane blockage. Such samples are crucial for teaching the model to represent and reason about rare, safety-critical edge cases.
\end{itemize}

\subsection{Hierarchical Generation}
\label{sec:supp_hierarchical_examples}

\begin{figure}[t]
    \centering
    \includegraphics[width=\linewidth]{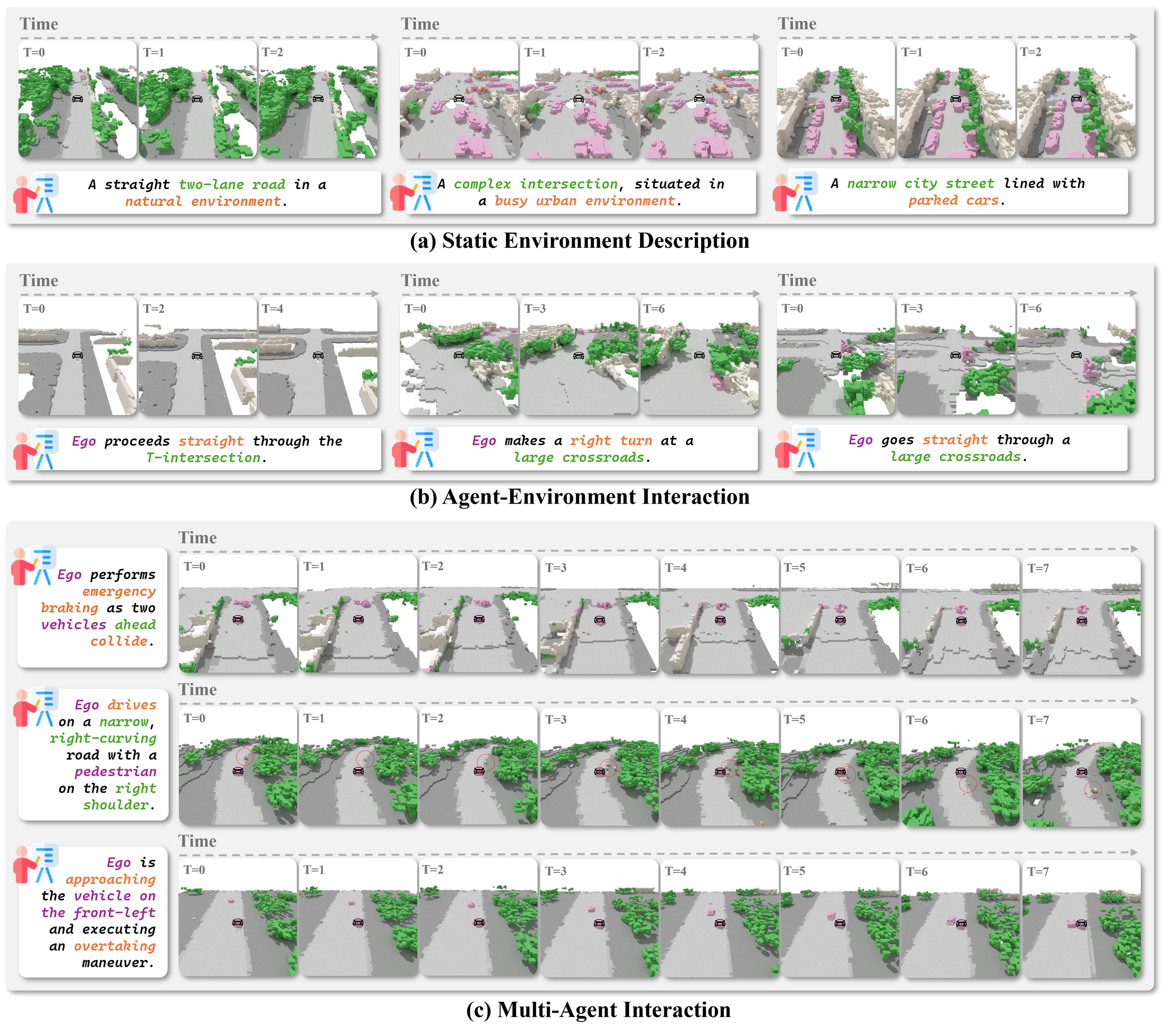}
    \caption{\textbf{Additional results on hierarchical generation control.} We visualize (a) static layout synthesis, (b) behavior-conditioned ego dynamics, and (c) multi-agent reactive orchestration.}
    \label{fig:supp_hierarchical_control}
\end{figure}

\cref{fig:supp_hierarchical_control} showcases the precision and diversity of \ModelName{} across three levels of hierarchical control, highlighting its superior semantic-to-spatial mapping.

\begin{enumerate}[label=(\alph*)]
    \item \textbf{Static Environment Description:} The model translates abstract descriptors (e.g., ``natural environment'' vs. ``busy urban intersection'') into geometrically distinct road topologies. Note how the model preserves fine-grained road boundaries and parked vehicle distributions, maintaining high-fidelity layout priors independent of temporal dynamics.
    \item \textbf{Agent-Environment Interaction:} \ModelName{} generates affordance-aware ego-trajectories that are strictly grounded in the generated layout. Whether performing a right turn at a crossroads or proceeding through a T-intersection, the ego's 4D volume remains logically aligned with the drivable road regions.
    \item \textbf{Multi-Agent Interaction:} The model orchestrates complex causal behaviors between the ego and other agents. In the emergency braking scenario, the model synthesizes a sudden obstacle and the corresponding rapid deceleration of the ego car, demonstrating an integrated understanding of dynamic agent logic and spatial constraints.
\end{enumerate}

\subsection{History-Prefix Control}
\label{sec:supp_more_history}

\begin{figure}[t]
    \centering
    \includegraphics[width=\linewidth]{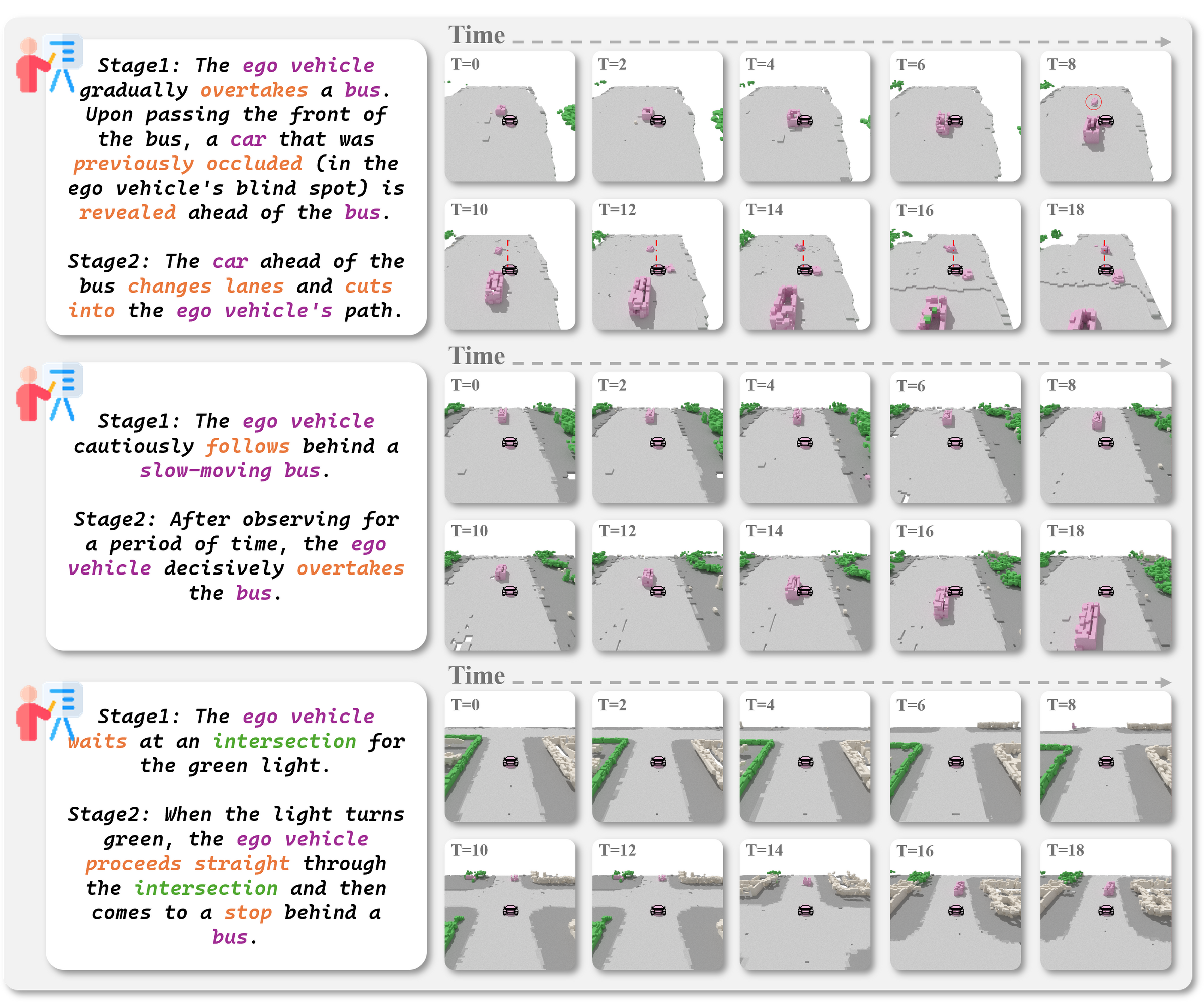}
    \caption{\textbf{Long-horizon narrative generation with history-prefix anchoring.} These multi-stage rollouts demonstrate the model's ability to maintain spatio-temporal coherence and respond to evolving instructions without losing scene consistency.}
    \label{fig:supp_long_horizon}
\end{figure}

To evaluate temporal consistency and causal logic over long horizons, we analyze multi-stage narratives in \cref{fig:supp_long_horizon}. These samples highlight how our history-prefix anchoring prevents ``scene-drifting'' and ensures logical transitions.

\begin{itemize}
    \item \textbf{Handling Occlusions and Transitions (Top):} The model demonstrates sophisticated spatial reasoning. As the ego overtakes a bus in Stage 1, a previously occluded car is revealed (marked with a \textit{red circle}). In Stage 2, the model smoothly transitions this revealed car into a lateral cut-in hazard. This confirms that \ModelName{} maintains a persistent internal representation of the scene, even for objects that are partially hidden or suddenly visible.
    \item \textbf{Persistent Scene Context (Middle):} During the decisive overtaking maneuver, the model maintains perfect spatial continuity of the bus and the background road layout across stages. The new instruction (``decisively overtake'') is seamlessly integrated with the historical context without altering the pre-established environment.
    \item \textbf{Grounding Procedural Logic (Bottom):} This complex stop-and-go sequence (waiting for a light $\to$ proceeding $\to$ stopping behind a bus) illustrates the model's grasp of procedural semantics. Unlike standard text-to-video models that often struggle with temporal ordering, \ModelName{} executes these logically dependent stages while maintaining a single, consistent 4D occupancy volume.
\end{itemize}

\subsection{Fine-grained Text-to-Occupancy Alignment}
\label{sec:supp_fine_grained}

\begin{figure}[t]
    \centering
    \includegraphics[width=\linewidth]{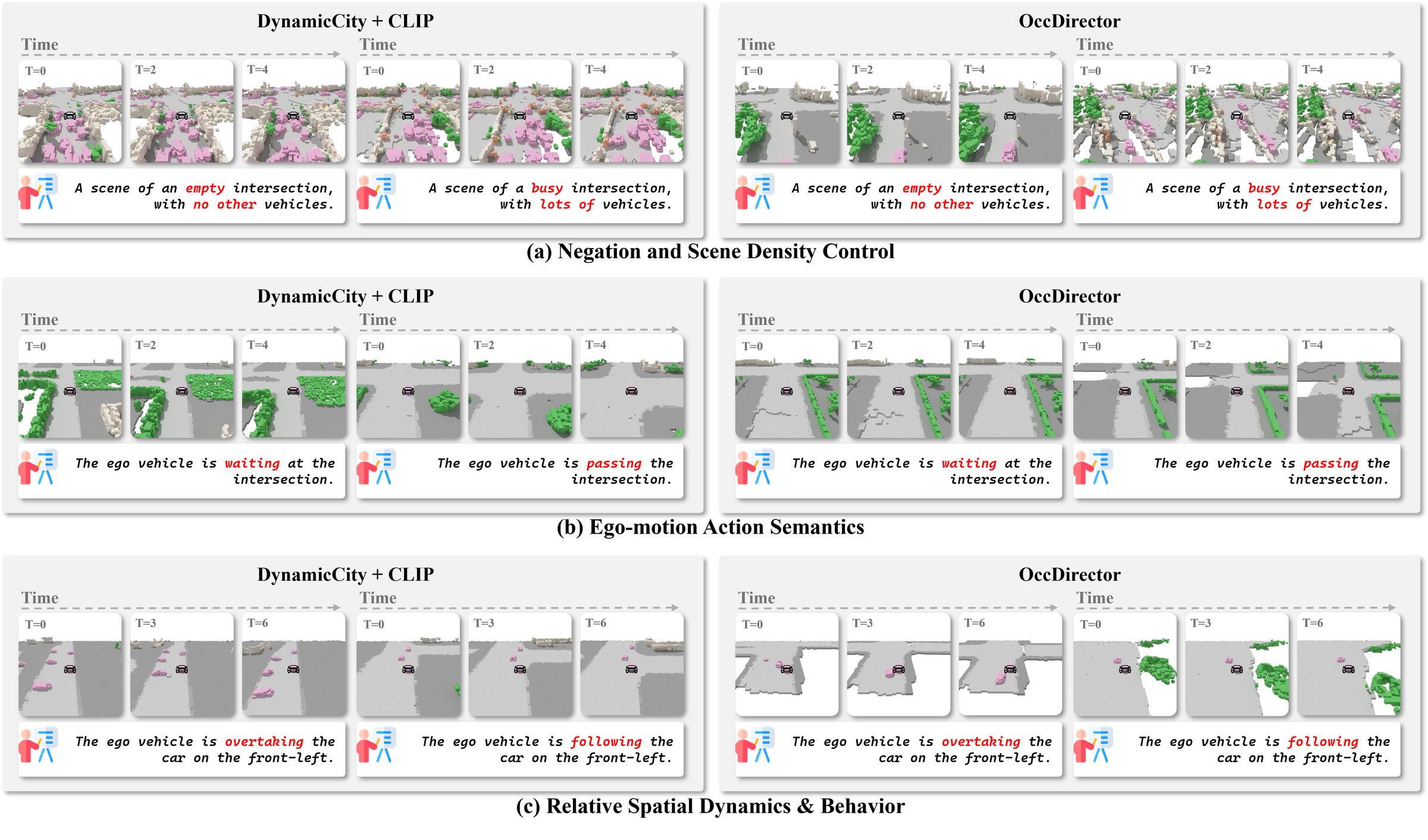}
    \caption{\textbf{Additional qualitative comparisons on fine-grained text alignment.} We showcase \ModelName{}'s ability to handle (a) negation and scene density, (b) ego-motion action semantics, and (c) complex relative spatial dynamics between agents.}
    \label{fig:supp_fine_grained_alignment}
\end{figure}

We provide additional qualitative comparisons in \cref{fig:supp_fine_grained_alignment} to further demonstrate \ModelName{}'s superior text-to-occupancy alignment compared to the DynamicCity + CLIP baseline.

\begin{enumerate}[label=(\alph*)]
    \item \textbf{Negation and Scene Density Control:} We evaluate the model's ability to process negative constraints (e.g., ``no other vehicles'') and control scene density. While \ModelName{} faithfully renders the requested environment, the baseline (DynamicCity + CLIP) fails to respect the negation, consistently hallucinating busy traffic in an ``empty intersection''.
    \item \textbf{Ego-motion Action Semantics:} The model correctly distinguishes between the ego vehicle ``waiting'' versus ``passing'' an intersection. In contrast, the baseline exhibits attribute confusion, depicting a moving ego vehicle even when the prompt explicitly requests it to wait, failing to decouple these distinct temporal states.
    \item \textbf{Relative Spatial Dynamics and Behavior:} \ModelName{} captures subtle interaction logic between agents. In the ``overtaking'' case, it synthesizes a precise lateral maneuver relative to a moving agent, whereas the baseline erroneously overtakes a static parked vehicle. For ``following'', \ModelName{} maintains stable longitudinal positioning, while the baseline fails to coordinate the relative spatial dynamics, generating extraneous vehicles that disrupt the intended interaction (\ie, additional car directly on the front).
\end{enumerate}

\section{Potential Societal Impact \& Limitations}
\label{sec:supp_impact}

\subsection{Potential Societal Impact}
\label{sec:supp_societal_impact}

Our work, \ModelName{}, contributes to the development of safer and more reliable autonomous driving (AD) systems by providing a high-fidelity, controllable generative model for 4D occupancy. By enabling the synthesis of diverse, safety-critical traffic scenarios from simple natural language prompts, it allows AD systems to be rigorously tested and validated against rare, high-risk edge cases that are difficult and dangerous to capture in the real world. This democratizes access to high-quality 4D training data for research institutions that may not have the resources for large-scale data collection.

However, like any generative technology, \ModelName{} could be misused. The ability to create realistic 4D traffic scenarios might be exploited to generate deceptive or ``deepfake'' data, potentially misleading safety evaluations or creating unrealistic performance benchmarks. Furthermore, an over-reliance on synthetic data could lead to models that fail to generalize to the full complexity of real-world environments, particularly if the generated distribution has subtle biases. We advocate for the responsible use of synthetic data as a supplement to, rather than a replacement for, real-world validation.

\subsection{Limitations}
\label{sec:supp_limitations}

Despite the advancements in \ModelName{}, several limitations remain:
\begin{itemize}
    \item \textbf{Data Artifacts:} Since \ModelName{} is trained on real-world logs (e.g., nuScenes, Waymo), it inherits inherent sensor artifacts, such as blind spots and missing ego-vehicle voxels, which may be reflected in the generated occupancy grids.
    \item \textbf{Long-horizon Consistency:} While history-prefix anchoring significantly improves temporal stability, generating extremely long sequences (beyond 8s) still faces challenges in maintaining absolute physical consistency due to the accumulation of small errors over time.
    \item \textbf{VLM-Judge Bias:} Our evaluation relies on rubric-based VLM judges (e.g., GPT-5.2, Gemini-3.0-pro). These models may exhibit subjective biases or differing calibration scales (e.g., GPT-5.2 being noticeably stricter than Gemini-3.0-pro), which can affect the absolute scores in human-like semantic assessment.
    \item \textbf{Semantic Granularity:} Our current model operates on 11 semantic categories. It does not yet capture fine-grained object states, such as vehicle turn signals, brake lights, or subtle pedestrian gestures, which are crucial for high-level interaction understanding.
\end{itemize}

\section{Public Resources Used}
\label{sec:supp_resources}

In this section, we acknowledge the public datasets, models, and open-source libraries used in this research.

\subsection{Data Source \& Privacy Declaration}
\textbf{Declaration on Data Usage and Privacy:} All data utilized in this study are derived strictly from publicly available open-source benchmarks and purely synthetic/simulation environments. No proprietary commercial data, vehicle onboard telemetry, or private real-world driving data from Li Auto or any other entity were used or accessed in this research.

\begin{itemize}
    \item UniOcc\footnote{\url{https://github.com/tasl-lab/uniocc}} \dotfill MIT License
\end{itemize}

\subsection{Public Models \& Benchmarking Vision-Language Models}
\textbf{Declaration on Model Usage:} The third-party proprietary and open-source Vision-Language Models (VLMs) and foundation models listed below are employed solely as public academic benchmarking baselines and data annotation tools.

\begin{itemize}
    \item Qwen3-VL\footnote{\url{https://github.com/QwenLM/Qwen3-VL}} \dotfill Apache 2.0
    \item OpenAI CLIP\footnote{\url{https://github.com/openai/CLIP}} \dotfill MIT License
    \item OpenAI GPT-5.2\footnote{\url{https://openai.com}} \dotfill Proprietary
    \item Google T5\footnote{\url{https://github.com/google-research/text-to-text-transfer-transformer}} \dotfill Apache 2.0
    \item Google Gemini-3.0-pro\footnote{\url{https://deepmind.google/technologies/gemini}} \dotfill Proprietary
\end{itemize}

\subsection{Public Implementations \& Libraries}
\textbf{Declaration on Code and Software Usage:} The open-source implementations, simulation tools, and third-party libraries listed below were utilized strictly for academic experimentation, validation, and paper implementation within this research project. None of these external codebases or dependencies are integrated into or affiliated with Li Auto's commercial software stack or production vehicle systems.

\begin{itemize}
    \item CarlaSC\footnote{\url{https://github.com/UMich-CURLY/3DMapping}} \dotfill MIT License
    \item CARLA Simulator\footnote{\url{https://github.com/carla-simulator/carla}} \dotfill MIT License
    \item TTSG\footnote{\url{https://github.com/basiclab/TTSG}} \dotfill No explicit license
    \item PyTorch Lightning\footnote{\url{https://github.com/Lightning-AI/pytorch-lightning}} \dotfill Apache 2.0
    \item DOME\footnote{\url{https://github.com/gusongen/DOME}} \dotfill Apache 2.0
    \item COME\footnote{\url{https://github.com/synsin0/COME}} \dotfill Apache 2.0
    \item DynamicCity\footnote{\url{https://github.com/3DTopia/DynamicCity}} \dotfill No explicit license
\end{itemize}

\setcounter{enumiv}{0}

\end{document}